\documentclass[journal,twoside,web]{ieeecolor2}
\usepackage{generic}
\usepackage{cite}
\usepackage{amsmath,amssymb,amsfonts}
\usepackage{algorithmic}
\usepackage{graphicx}
\usepackage{textcomp}

\usepackage{hyperref}
\hypersetup{hypertex=true,
colorlinks=true,
linkcolor=blue,
anchorcolor=blue,
citecolor=blue}
\usepackage{multicol}
\usepackage{multirow}
\usepackage{booktabs}
\usepackage[table]{xcolor}
\definecolor{deepgreen}{RGB}{0,180,0}
\definecolor{lightblue}{RGB}{221, 242, 255}
\usepackage{bm}
\usepackage{wrapfig}
\usepackage{subcaption}
\usepackage{soul}
\usepackage{tabularx}
\usepackage{array}

\def\BibTeX{{\rm B\kern-.05em{\sc i\kern-.025em b}\kern-.08em
    T\kern-.1667em\lower.7ex\hbox{E}\kern-.125emX}}
\begin{document}

\title{Vascular Anatomy-aware Self-supervised Pre-training for X-ray Angiogram Analysis}

\author{
De-Xing Huang, 
Chaohui Yu, 
Xiao-Hu Zhou,~\IEEEmembership{Member,~IEEE}, 
Tian-Yu Xiang, 
Qin-Yi Zhang, 
Mei-Jiang Gui, 
Rui-Ze Ma, 
Chen-Yu Wang, 
Nu-Fang Xiao, 
Fan Wang, and 
Zeng-Guang Hou,~\IEEEmembership{Fellow,~IEEE}
\thanks{This work was supported in part by the National Key Research and Development Program of China under Grant 2023YFC2415100, in part by the National Natural Science Foundation of China under Grant 62633021, Grant 62373351, Grant 62503474, Grant 82327801, Grant 62303463, Grant 52475038, in part by the Chinese Academy of Sciences Project for Research Instrument Development under Grant No. PTYQ2026YZ0072, Young Scientists in Basic Research under Grant No. YSBR-104, in part by the Beijing Natural Science Foundation under Grant F252068, Grant 4254107, in part by Beijing Nova Program under Grant 20250484813, in part by China Postdoctoral Science Foundation under Grant 2024M763535, in part by the Postdoctoral Fellowship Program of CPSF under Grant GZC20251170, and in part by Xiaomi Young Talents Program/Xiaomi Foundation. (Corresponding authors: \textit{Xiao-Hu~Zhou} and \textit{Zeng-Guang~Hou})}
\thanks{D.-X. Huang, X.-H. Zhou, T.-Y. Xiang, Q.-Y. Zhang, M.-J. Gui, R.-Z. Ma, C.-Y. Wang, N.-F. Xiao and Z.-G. Hou are with the State Key Laboratory of Multimodal Artificial Intelligence Systems, Institute of Automation, Chinese Academy of Sciences, Beijing 100190, China (email: {\tt \{\href{mailto:huangdexing2022@ia.ac.cn}{huangdexing2022}, \href{mailto:xiaohu.zhou@ia.ac.cn}{xiaohu.zhou}, \href{mailto:zengguang.hou@ia.ac.cn}{zengguang.hou}\}@ia.ac.cn}).}
\thanks{D.-X. Huang, X.-H. Zhou, T.-Y. Xiang, Q.-Y. Zhang and Z.-G. Hou are also with the School of Artificial Intelligence, University of Chinese Academy of Sciences, Beijing 100049, China.}
\thanks{C. Yu and F. Wang are with the DAMO Academy, Alibaba Group, Hangzhou 310023, China. C. Yu is also with Hupan Lab, Hangzhou 310023, China.}
}

\maketitle

\begin{abstract}
\textit{Objective}: X-ray angiography is the gold standard imaging modality for cardiovascular diseases. However, current deep learning approaches for X-ray angiogram analysis are severely constrained by the scarcity of annotated data. While large-scale self-supervised learning (SSL) has emerged as a promising solution, its potential in this domain remains largely unexplored, primarily due to the lack of \textit{effective SSL frameworks} and \textit{large-scale datasets}. \textit{Methods}: To bridge this gap, we introduce a \underline{\textbf{Vas}}cular anat\underline{\textbf{o}}my-aware \underline{\textbf{M}}asked \underline{\textbf{I}}mage \underline{\textbf{M}}odeling (VasoMIM) framework that explicitly integrates domain-specific anatomical knowledge. Specifically, VasoMIM comprises two key designs: an \textit{anatomy-guided masking strategy} and an \textit{anatomical consistency loss}. The former strategically masks vessel-containing patches to compel the model to learn robust vascular semantics, while the latter preserves structural consistency of vessels between the original and reconstructed images, enhancing the discriminability of the learned representations. In conjunction with VasoMIM, we curate XA-170K, the largest X-ray angiogram pre-training dataset to date. \textit{Results}: We validate VasoMIM on four downstream tasks across six datasets, where it demonstrates superior transferability and achieves state-of-the-art performance compared to existing methods. \textit{Conclusion}: Incorporating vascular anatomical knowledge into self-supervised pre-training enables more discriminative and transferable representations for X-ray angiogram analysis. \textit{Significance}: These findings highlight the significant potential of VasoMIM as a foundation model for advancing a wide range of X-ray angiogram analysis tasks. VasoMIM and XA-170K are available on \href{https://github.com/Dxhuang-CASIA/XA-SSL}{GitHub} and \href{https://huggingface.co/datasets/waha2000huang/XA-170K}{Hugging Face}, respectively.
\end{abstract}


\section{Introduction} \label{sec:introduction}
\IEEEPARstart{C}{ardiovascular} diseases (CVDs) constitute a global health crisis and remain the leading cause of mortality worldwide~\cite{vaduganathan2022global}. X-ray angiography serves as the gold standard for diagnosing CVDs~\cite{kheiri2022computed}, planning interventions~\cite{writing20222021}, and guiding procedures~\cite{huang2025real}. However, accurate interpretation of these angiograms is fraught with challenges. Radiologists often struggle to precisely delineate vascular topology or identify stenosis due to low contrast, motion artifacts, and overlapping anatomical structures~\cite{huang2026spironet}. Consequently, there is an urgent demand for automated, robust image analysis methodologies.

With the advent of artificial intelligence (AI), numerous techniques have been proposed to reduce clinical workload, ranging from vessel segmentation~\cite{padariya2025novel, huang2026spironet} to stenosis detection~\cite{li2024stqd,chen2025hierarchical}. Despite these advances, training high-performance models typically requires task-specific annotations, such as pixel-level masks for segmentation and bounding boxes for detection. Producing such annotations is labor-intensive, time-consuming, and dependent on specialized domain knowledge~\cite{esteva2019guide}. To address this bottleneck, the community is shifting from supervised training on small datasets to self-supervised learning (SSL), which leverages vast amounts of unlabeled data to learn generalizable representations~\cite{gui2024survey}. Yet, the potential of SSL in this domain remains largely underexplored. This gap can be attributed to two primary hurdles: the lack of \textit{\textbf{domain-specific SSL frameworks}} and the absence of \textit{\textbf{large-scale datasets}}.

\begin{figure}[htbp]
\centering\centerline{\includegraphics{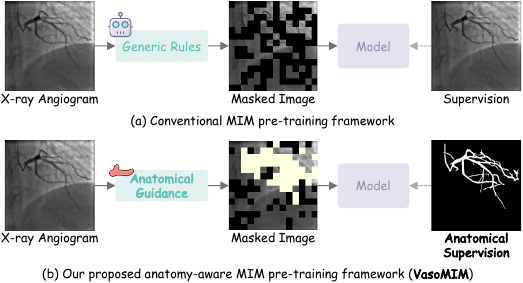}}
\caption{Conceptual comparison. (a) Conventional MIM masks patches based on \textit{generic rules} and utilizes \textit{standard supervision} to reconstruct the original image. (b) VasoMIM employs \textit{anatomical guidance} to selectively mask vessel-relevant regions (highlighted in yellow) and enforces structural consistency via \textit{anatomical supervision}, enabling the model to learn richer vascular representations.}
\label{fig:conceptual}
\end{figure}
\begin{figure}[htbp]
\centering\centerline{\includegraphics{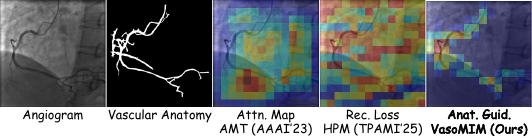}}
\caption{Generic \textit{vs.} anatomy-guided masking strategies. While generic methods like AMT~\cite{liu2023good} and HPM~\cite{wang2025bootstrap} fail to prioritize vascular structures, VasoMIM effectively focuses on vessel-relevant regions. Red denotes a higher masking probability, while blue indicates the opposite.}
\label{fig:guidance_compare}
\end{figure}

\begin{table*}[t]
\caption{Comparison between the preliminary conference version VasoMIM-v1~\cite{huang2026vasomim} and this extended version.}
\label{table:comparison_aaai}
\centering
\small
\setlength{\tabcolsep}{4pt}
\renewcommand{\arraystretch}{1.25}
\begin{tabularx}{\linewidth}{
    >{\raggedright\arraybackslash}p{0.16\linewidth}
    >{\raggedright\arraybackslash}p{0.25\linewidth}
    >{\raggedright\arraybackslash}p{0.25\linewidth}
    >{\raggedright\arraybackslash}X
}
\toprule
Aspect & VasoMIM-v1~\cite{huang2026vasomim} & VasoMIM & Main Contribution \\ 
\midrule
Anatomical Guidance  
& Instance-level guidance from Frangi filter 
& Incorporating population-level guidance learned from large-scale X-ray angiograms 
& More reliable anatomical guidance by complementing handcrafted vesselness cues with learned anatomical knowledge \\ 
\midrule

Pre-training Dataset 
& $20$K X-ray angiograms 
& $170$K X-ray angiograms 
& The largest publicly available dataset for X-ray angiogram pre-training \\ 
\midrule

Downstream Validation 
& $1$ task (vessel segmentation) on $3$ datasets 
& $4$ tasks (segmentation and detection) on $6$ datasets 
& More comprehensive evaluation across diverse X-ray angiogram analysis tasks \\ 
\midrule

Experimental Analysis 
& Basic ablation studies 
& Extended ablations and analyses of scalability, label efficiency, pre-training datasets, and robustness 
& More in-depth understanding of the effectiveness and generalizability of VasoMIM \\ 
\bottomrule
\end{tabularx}
\end{table*}

As a representative SSL technique, masked image modeling (MIM) has achieved remarkable success in medical image analysis~\cite{he2022masked,zhuang2025mim,tang2025mambamim} by training models to reconstruct masked image patches (Fig.~\ref{fig:conceptual} (a)). However, the direct application of standard MIM to X-ray angiograms is suboptimal due to the \textit{extreme sparsity of vascular structures}: \textbf{\textit{i)} Anatomically agnostic masking strategies fail to prioritize sparse vessel-relevant regions.} Current strategies rely on generic rules, either data-independent (\textit{e.g.}, random~\cite{he2022masked}, block-wise~\cite{bao2022beit}) or data-adaptive (\textit{e.g.}, attention-based~\cite{liu2023good} or loss-based~\cite{wang2025bootstrap}). While effective for natural images with dense semantic content, these strategies often ignore thin vascular structures. As shown in Fig.~\ref{fig:guidance_compare}, standard strategies select a large proportion of background-only patches simply because they dominate image statistics, rendering the model insensitive to vascular anatomy. \textbf{\textit{ii)} Conventional reconstruction objectives lack semantic discriminability.} Existing approaches learn reconstruction via pixel-level regression~\cite{he2022masked} or feature distillation~\cite{zhou2022ibot}. Pixel-level loss encourages the prediction of low-frequency background textures rather than high-frequency vascular details. While feature distillation methods~\cite{bao2022beit,wang2025bootstrap} attempt to alleviate this, they are computationally expensive and, being pre-trained on natural images, lack domain-specific semantics.

To address these challenges, we introduce \textbf{\underline{Vas}}cular anat\textbf{\underline{o}}my-aware \textbf{\underline{M}}asked \textbf{\underline{I}}mage \textbf{\underline{M}}odeling (\textbf{VasoMIM}), a framework tailored for X-ray angiograms, as presented in Fig.~\ref{fig:conceptual} (b). The core insight is to inject a strong anatomical inductive bias into the model. \textbf{First, we design an \textit{anatomy-guided masking strategy} that directs the model's focus toward vessel-relevant regions.} In our preliminary version~\cite{huang2026vasomim}, we directly adopt vascular anatomy extracted by Frangi filter~\cite{frangi1998multiscale} as guidance, which provides instance-specific anatomical knowledge for identifying vessel-relevant regions. However, such handcrafted guidance is based on local vesselness responses and can be affected by noise, low contrast, or non-vascular structures. This limitation becomes more evident when scaling to a larger pre-training dataset. In this work, we further introduce a probability map produced by a lightweight segmentor (UNeXt-S~\cite{valanarasu2022unext}, $0.26$ M), which encodes population-level anatomical knowledge learned from large-scale X-ray angiograms. The resulting co-guidance strategy combines image-specific vesselness cues with learned vascular semantics, enabling more reliable localization of informative vessel-relevant regions (Fig.~\ref{fig:guidance_compare}). \textbf{Second, we introduce an \textit{anatomical consistency loss} to enhance representation discriminability.} Specifically, we enforce consistency between the vessel segmentation of the original and reconstructed images using the lightweight segmentor, encouraging the model to learn topologically accurate representations rather than merely reconstructing pixel intensities.

Effective pre-training requires data at scale. Unlike other medical imaging modalities such as CT~\cite{wu2026large}, MRI~\cite{qiu2025large}, and chest X-ray~\cite{perez2025exploring}, where pre-training datasets have reached hundreds of thousands of samples, the X-ray angiogram domain still lacks established large-scale datasets. In our preliminary version~\cite{huang2026vasomim}, we curated a pre-training dataset comprising $20$K X-ray angiograms. Although this dataset enabled an initial validation of anatomy-aware pre-training, its relatively limited scale may not sufficiently cover the substantial variations in vascular anatomy. Such limited data diversity may restrict the robustness and transferability of the learned representations. In this study, we substantially scale up the dataset and curate XA-$170$K. To the best of our knowledge, XA-$170$K is the largest publicly available dataset for X-ray angiogram pre-training.

The main contributions of this study are summarized as follows:
\begin{itemize}
    \item A novel self-supervised pre-training framework, VasoMIM, is proposed for X-ray angiogram analysis. This method explicitly integrates vascular anatomical knowledge to drive robust vascular representation learning.
    \item Two vascular anatomy-aware designs are proposed: \textit{\textbf{i)}} an anatomy-guided masking strategy that preferentially masks vessel-containing patches, compelling the model to focus on learning robust vascular representations; and \textit{\textbf{ii)}} an anatomical consistency loss that preserves the structural consistency of vessels between the original and reconstructed images, thereby enhancing the discriminability of the learned representations.
    \item The largest X-ray angiogram pre-training dataset to date, XA-$170$K, is introduced. Leveraging this dataset, we provide the first systematic empirical analysis of data and model scaling for MIM in this specific domain.
    \item Extensive experiments across four downstream tasks and six datasets demonstrate that VasoMIM significantly outperforms the state-of-the-art SSL alternatives.
\end{itemize}

The preliminary version of this study was published at \textit{AAAI} 2026~\cite{huang2026vasomim}. In this extended version, we present substantial new contributions, as summarized in Table~\ref{table:comparison_aaai}: \textbf{\textit{i)} Anatomy-aware Co-guidance.} Compared to VasoMIM-v1~\cite{huang2026vasomim}, which relied solely on vascular anatomy extracted by Frangi filter as instance-specific guidance, we further incorporate segmentor-derived probability maps as population-level anatomical knowledge. This co-guidance design combines image-specific vesselness responses with learned vascular semantics from large-scale X-ray angiograms, thereby providing more robust guidance for localizing vessel-relevant regions. \textbf{\textit{ii)} Scaling Up the Pre-training Data.} We expand the pre-training dataset from $20$K to $170$K. Crucially, we systematically verify the scalability of VasoMIM regarding both data scale and model capacity, providing empirical evidence for the benefits of large-scale pre-training. \textbf{\textit{iii)} Comprehensive Evaluation.} We expand validation from a single task to four distinct clinical applications: vessel segmentation, vessel segment segmentation, stenosis segmentation, and stenosis detection. Furthermore, we include a wider range of state-of-the-art baselines to ensure a rigorous and comprehensive comparison. \textbf{\textit{iv)} In-depth Analysis.} We provide detailed analyses of the core design choices, offering deeper insights into the mechanism of anatomy-aware pre-training.

The remainder of this paper is organized as follows: Section~\ref{sec:related_works} briefly reviews some techniques related to this research. Section~\ref{sec:method} describes the proposed VasoMIM in detail. Section~\ref{sec:results} presents the experimental configurations and results. An in-depth analysis of VasoMIM is provided in Section~\ref{sec:analysis}. Section~\ref{sec:limitation} discusses the current limitations and outlines potential directions for future works. Finally, Section~\ref{sec:conclusion} concludes this paper.

\section{Related Works} \label{sec:related_works}
\subsection{Self-supervised Medical Image Pre-training}
Self-supervised learning (SSL) can learn transferable visual representations without annotations~\cite{chen2021empirical,caron2021emerging,bao2022beit,xie2022simmim,he2022masked,maghsoodi2026domain,zhang2025bdfm}. It has become a cornerstone for large-scale vision pre-training~\cite{gui2024survey}. State-of-the-art SSL approaches generally fall into two categories: contrastive learning (CL) and masked image modeling (MIM). CL methods~\cite{zhou2020comparing,chen2021empirical,wu2026large} aim to align positive pairs while separating negative pairs in the embedding space. While effective for image-level tasks like classification, these methods often struggle with dense prediction tasks~\cite{chaitanya2020contrastive}, such as segmentation and detection, which require fine-grained spatial representations. Conversely, MIM approaches train models to reconstruct masked patches from visible context, enabling the learning of detailed local representations~\cite{yuan2023hap}. Generic MIM frameworks like MAE~\cite{he2022masked} and SimMIM~\cite{xie2022simmim} have been successfully adapted to medical imaging, with further enhancements tailored to specific modalities including ultrasound~\cite{kang2024deblurring}, CT~\cite{zhuang2025mim}, fMRI~\cite{wang2025towards}, and pathology~\cite{xu2024whole}. However, directly applying existing SSL methods to X-ray angiograms remains suboptimal. Unlike many natural images with relatively dense semantic content, X-ray angiograms contain extremely sparse and thin vascular structures that can be easily overwhelmed by dominant background regions. As a result, generic SSL methods may fail to effectively learn vascular representations. Moreover, models pre-trained on natural images or other medical modalities suffer from domain gaps when transferred to X-ray angiogram analysis. These limitations highlight the need for both a domain-specific SSL framework that explicitly incorporates vascular anatomical knowledge and a large-scale X-ray angiogram dataset for pre-training.

\begin{figure*}[t]
\centering\centerline{\includegraphics{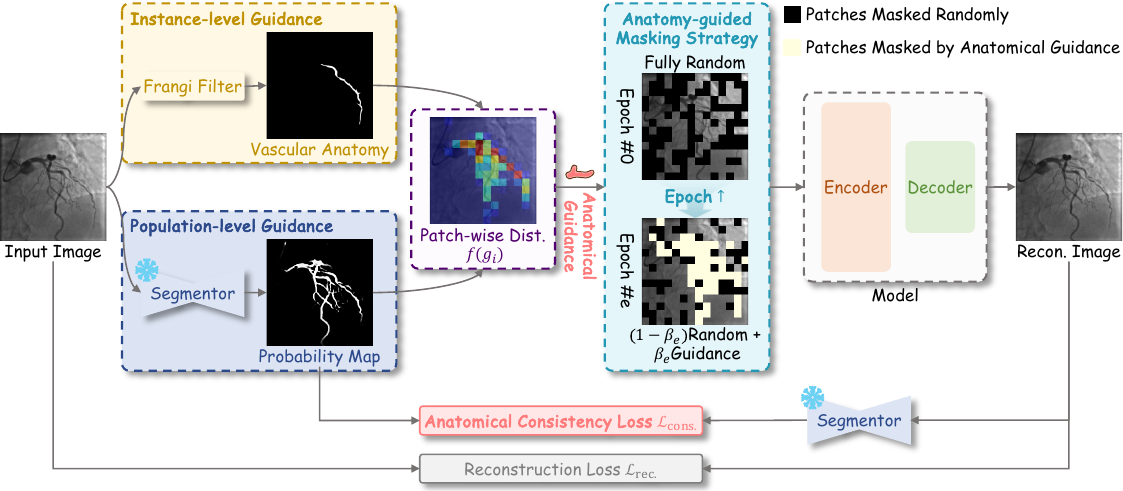}}
\caption{Overview of VasoMIM. Given an input X-ray angiogram, VasoMIM constructs anatomical guidance from two complementary sources: instance-level vesselness cues extracted by Frangi filter and population-level anatomical knowledge learned by a lightweight segmentor from the large-scale XA-$170$K. Based on the co-guidance from these two sources, a patch-wise vascular anatomical distribution $f(g_i)$ is computed to guide the masking strategy, thereby prioritizing vessel-relevant regions during pre-training. The masked image is then reconstructed by an encoder-decoder model. Finally, the model is optimized by minimizing the total objective $\mathcal{L}_{\rm MIM}$, which combines the standard pixel-level reconstruction loss $\mathcal{L}_{\rm rec.}$ and the anatomical consistency loss $\mathcal{L}_{\rm cons.}$ to learn discriminative vascular representations.}
\label{fig:vasomim}
\end{figure*}

\subsection{Masking Strategies and Reconstruction Objectives in Masked Image Modeling}
\textbf{Masking Strategies.} The effectiveness of MIM relies heavily on the masking strategy, particularly given the high spatial redundancy of images~\cite{he2022masked}. While simple random masking with high ratios has become the standard practice~\cite{kang2024deblurring,wang2025bootstrap}, it often fails to prioritize the most informative regions. Beyond random masking, various carefully designed masking strategies have been proposed to improve patch selection. For example, AMT~\cite{liu2023good} uses attention maps to guide masking, while HAP~\cite{yuan2023hap} incorporates human-structure priors for human-centric perception. Other approaches, such as HPM~\cite{wang2025bootstrap}, AnatoMask~\cite{li2024anatomask}, and AHM~\cite{xu2025self}, employ loss predictors, self-distillation, or policy networks to identify and mask ``hard'' patches. Despite their effectiveness in generic scenarios, these methods remain insensitive to the sparse vascular structures in X-ray angiograms due to the extreme dominance of the background, as illustrated in Fig.~\ref{fig:guidance_compare}. Our preliminary VasoMIM-v1~\cite{huang2026vasomim} addressed this issue by introducing vascular anatomy extracted by Frangi filter as guidance. However, such guidance mainly reflects instance-specific vesselness cues derived from each individual X-ray angiogram and is therefore susceptible to noise interference. This may lead to incomplete or noisy localization of vessel-relevant regions, limiting the learning of robust vascular representations. In this work, we further introduce a learned anatomical probability map that encodes population-level anatomical knowledge from large-scale X-ray angiograms to complement the instance-specific knowledge, thereby providing more reliable guidance for localizing informative vessel-relevant regions.

\textbf{Reconstruction Objectives.} MIM trains models to reconstruct masked patches according to predefined objectives. The most common reconstruction objectives minimize pixel-level errors, such as mean squared error (MSE)~\cite{he2022masked} or mean absolute error (L$_1$)~\cite{xie2022simmim}. However, pixel-level objectives mainly encourage low-level appearance reconstruction and can be dominated by background regions, especially in X-ray angiograms where vascular pixels are sparse and highly imbalanced with background pixels. As a result, they may not sufficiently preserve high-level vascular anatomical semantics. Alternative objectives distill high-level representations from pre-trained foundation models~\cite{bao2022beit,wang2025bootstrap}. While effective for natural images, such generic models may not fully capture the domain-specific vascular semantics of X-ray angiograms. To provide more suitable anatomical supervision, we leverage Frangi-derived pseudo-labels~\cite{frangi1998multiscale} to train a lightweight segmentor~\cite{valanarasu2022unext} ($0.26$ M). This segmentor serves as a domain-specific semantic extractor, enabling anatomical consistency supervision during MIM pre-training without relying on foundation models pre-trained on large-scale external data.

\subsection{X-ray Angiogram Analysis}
In clinical practice, X-ray angiography is the primary modality for visualizing vascular morphology and localizing stenosis~\cite{writing20222021}. Existing deep learning methods for X-ray angiogram analysis are typically trained from scratch under fully supervised settings, often adopting architectures such as UNet~\cite{ronneberger2015u} for segmentation and Faster R-CNN~\cite{ren2016faster} for detection. Subsequent studies have improved these frameworks through attention mechanisms~\cite{li2020cau,zhang2020direct}, multi-scale fusion~\cite{huang2026spironet,chen2025hierarchical}, and advanced network architectures~\cite{chen2024transunet,ruan2025vm,jiang2025rwkv,li2024stqd}. However, these models usually learn task-specific representations, which may generalize poorly to novel tasks~\cite{moor2023foundation}. Moreover, the annotations required for training are expensive to obtain in X-ray angiograms. In contrast to these approaches, we develop a domain-specific self-supervised pre-training framework for X-ray angiogram analysis using large-scale unlabeled data, aiming to learn robust representations that can transfer effectively across diverse downstream tasks.

\section{Method} \label{sec:method}
Fig.~\ref{fig:vasomim} summarizes the overall pipeline. Section~\ref{sec:method_frangi} first describes how Frangi filter produces an instance-specific vascular mask. Section~\ref{sec:method_masking} then explains how this mask is combined with a population-level probability map to guide patch selection. Finally, Section~\ref{sec:method_loss} presents the anatomical consistency objective used to preserve vascular structure during reconstruction.

\subsection{Frangi Filter as the Anatomical Extractor}~\label{sec:method_frangi}
We leverage the Hessian-based Frangi filter to extract vascular anatomy in an unsupervised manner. Following the implementation in~\cite{wu2025denver}, the procedure consists of three stages.

\textbf{Multi-scale Hessian Analysis.} Let $I\in \mathbb{R}^{C\times H\times W}$ denote the input X-ray angiogram. To capture vessels of varying calibers, we analyze the local curvature of image intensity by convolving $I$ with Gaussian derivative kernels $G(\sigma)$ across a set of scales $\sigma\in\left\{1,2,3,4\right\}$. The Hessian matrix at the $i$-th pixel is computed as $H_\sigma(i)=\nabla^2\left[I * G(\sigma)\right]\left(i\right)$. By analyzing the eigenvalues $\left|\lambda_1\right|\leq\left|\lambda_2\right|$ of $H_\sigma(i)$, vascular structures are enhanced while suppressing background noise. Specifically, the vesselness response $V_\sigma(i)$ is formulated as:
\begin{align}
V_{\sigma}(i) =
\begin{cases}
\lvert \lambda_{1}(i)\rvert, & \lambda_{2}(i) < 0,\\
0, & {\rm otherwise.}
\end{cases}
\end{align}

The aggregated vesselness map $V(i)$ is obtained by taking the maximum response across all considered scales, ensuring robust extraction of both thin and thick vessels.

\textbf{Percentile-based Adaptive Thresholding.} To convert the continuous vesselness map $V(i)$ into a discrete binary mask, we employ an adaptive thresholding strategy. Instead of a fixed intensity threshold, we calculate a dynamic threshold $T={\rm Percentile}_{\alpha}\left[V\left(i\right)\right]$ based on the $\alpha$-th percentile of the vesselness distribution. A preliminary binary map is generated via $\hat{B}\left(i\right)=\mathbb{I}\left[V(i)\geq T\right]$. This step effectively suppresses low-confidence noise while preserving salient vascular features. We set $\alpha=92$ by default.

\textbf{Region Growing Refinement.} In the final stage, we enforce spatial connectivity to eliminate isolated noise artifacts. We first identify the global maximum of the vesselness map to serve as a seed point $s=\arg\max_i V\left(i\right)$. Starting from $s$, a region-growing algorithm~\cite{adams1994seeded} iteratively merges neighboring pixels that exhibit high vesselness values. This process yields the final mask $B\in\left\{0,1\right\}^{1\times H\times W}$, which delineates the continuous vascular topology.

\subsection{Anatomy-guided Masking Strategy}~\label{sec:method_masking}
Compared to natural images, X-ray angiograms are characterized by dominant background regions and sparse vascular structures. Generic masking strategies lack anatomical awareness and therefore tend to select background-only patches. Consequently, pre-training is dominated by the reconstruction of non-informative background textures rather than the learning of vascular representations.

To overcome this bottleneck, we introduce an anatomy-guided masking strategy. Our core insight is that \textit{patches containing richer vascular structures carry more anatomical information and should be masked with higher probability}.

\textbf{Revisiting Instance-level Guidance in VasoMIM-v1~\cite{huang2026vasomim}.} In our preliminary work, anatomical guidance is derived from Frangi filter. Specifically, for each individual X-ray angiogram, we extract a binary mask $B\in\left\{0,1\right\}^{1\times H\times W}$ and use it to identify vessel-relevant patches. This instance-level guidance allows the masking strategy to focus more on vascular regions than generic masking strategies. However, Frangi filter relies on handcrafted local vesselness responses and processes each image independently. As a result, it can be affected by noise, low contrast, bony structures, and other non-vascular patterns, leading to incomplete or noisy vessel localization. Such image-specific errors can reduce the reliability of Frangi-derived guidance, particularly when pre-training is scaled to a large and diverse dataset.

\textbf{Population-level Guidance.} To make anatomical guidance more reliable under large-scale pre-training, we further introduce population-level anatomical knowledge learned from XA-$170$K. Specifically, we train a lightweight segmentor, \textit{i.e.}, UNeXt-S~\cite{valanarasu2022unext}, using pseudo-labels generated by Frangi filter. Unlike Frangi filter, which extracts vesselness independently from each angiogram, the segmentor is jointly trained across the entire XA-$170$K dataset and learns common vascular patterns from a large population of angiograms. Thus, although its supervision originates from Frangi-derived pseudo-labels, the learned segmentor can aggregate population-level anatomical information and reduce image-specific noise and fragmented responses. We use its predicted probability map $M\in\left[0,1\right]^{1\times H\times W}$ as learned anatomical guidance, which complements the instance-level Frangi response with smoother and more complete vascular information.

\textbf{Co-guidance.} Based on the above considerations, we combine the Frangi-derived instance-level guidance and the segmentor-derived population-level guidance to form a co-guidance map $G\in[0,1]^{1\times H\times W}$:
\begin{align}
G = \eta \cdot B + (1-\eta) \cdot M~\label{eq:mixing}
\end{align}
where $\eta$ controls the relative contribution of the two guidance sources. We set $\eta=0.5$ by default to balance the instance-level and population-level information.

During pre-training, the angiogram $I$ and the corresponding co-guidance map $G$ are partitioned into non-overlapping patches $x\in\mathbb{R}^{N\times(P^2\cdot C)}$ and $g\in\mathbb{R}^{N\times P^2}$, respectively, where $P$ is the patch size and $N=(H\cdot W)/P^2$ is the sequence length. Let $g_i\in[0,1]^{P^2}$ denote the $i$-th patch of $g$. The patch-wise vascular anatomical distribution is then defined as:
\begin{align}
    f(g_i)=\frac{\sum_{j=1}^{P^2} g_{ij}}{\sum_{k=1}^{N}\sum_{j=1}^{P^2} g_{kj}}, \quad i=1,2,\cdots,N
\end{align}

Consequently, $f(g_i)$ assigns higher sampling probabilities to patches with denser vascular responses, thereby prioritizing anatomically informative regions for masking. In practice, we sample patches from this distribution using {\tt torch.multinomial}.

\textbf{Weak-to-Strong Anatomical Guidance.} Masking an excessive number of vessel-containing patches in the early training stages may overwhelm the model, impeding convergence. To address this, we implement a weak-to-strong anatomy-guided strategy. As illustrated in Fig.~\ref{fig:vasomim}, for a specific epoch $e$, only a fraction $\beta_e$ of the masked patches is sampled according to $f(g_i)$, while the remaining fraction, $1-\beta_e$, is selected randomly, following standard MAE~\cite{he2022masked}. Thus, $\beta_e$ controls the strength of anatomical guidance during training. The guidance intensity $\beta_e$ increases linearly:
\begin{table}[htbp]
\caption{Statistics of XA-$170$K and downstream datasets.}
\label{table:dataset}
\centering
\begin{tabular}{llll}
\toprule
Usage & Dataset & \#train & \#test \\ \midrule
\multirow{5}{*}{Pre-training}  & CADICA~\cite{jimenez2024cadica} & $6,594$ & $-$ \\
 & SYNTAX~\cite{mahmoudi2025x} & $2,943$ & $-$ \\
 & XCAD~\cite{ma2021self} & $1,621$ & $-$ \\
 & CoronaryDominance~\cite{kruzhilov2025coronarydominance} & $160,320$ & $-$ \\ 
 & \cellcolor{lightblue}Total & \cellcolor{lightblue}$171,478$ & \cellcolor{lightblue}$-$ \\ 
 \midrule
\multirow{5}{*}{Segmentation} & ARCADE-V~\cite{popov2024dataset} & $1,000$ & $300$ \\
 & ARCADE-VS~\cite{popov2024dataset} & $1,000$ & $300$ \\
 & ARCADE-S~\cite{popov2024dataset} & $1,000$ & $300$ \\
 & CAXF~\cite{li2020cau} & $337$ & $201$ \\
 & XACV~\cite{wu2025denver} & $175$ & $46$ \\
 \midrule
Detection & Stenosis~\cite{danilov2021real} & $7,492$ & $833$ \\ \bottomrule
\end{tabular}
\end{table}
\begin{align}
    \beta_e=\beta_0+\frac{e}{E}\left(\beta_E-\beta_0\right)
\end{align}
where $E$ is the maximum pre-training epoch, and $\beta_0, \beta_E \in [0,1]$ are hyper-parameters. Given a masking ratio $\gamma$, for each input image at epoch $e$, a total of $\gamma N$ patches are masked at each epoch, among which $\beta_e\gamma N$ are sampled according to $f\left(g_i\right)$ and $(1-\beta_e)\gamma N$ are sampled randomly. This schedule gradually increases the influence of anatomical guidance while preserving sufficient random context for stable pre-training.

\subsection{Anatomical Consistency Loss}~\label{sec:method_loss}
Conventional MIM reconstruction objectives usually rely on pixel-level regression, which encourages low-level appearance reconstruction but does not guarantee the preservation of high-level vascular semantics. This limitation is particularly critical for X-ray angiogram analysis, where the continuity and topology of vascular structures are more important than exact pixel intensity reconstruction. Although feature distillation from pre-trained foundation models can provide semantic supervision, these generic teachers may suffer from domain gaps and fail to capture the vascular topology specific to X-ray angiograms. To bridge this gap, we introduce an anatomical consistency loss that explicitly enforces the preservation of vascular topology in the reconstructed image:
\begin{align}
    \mathcal{L}_{\rm cons.}=\mathcal{L}\left(\mathcal{S}(I), \mathcal{S}(I^\prime)\right)
\end{align}
where $I^\prime$ is the reconstructed angiogram, and $\mathcal{S}(\cdot)$ is the lightweight segmentor introduced in Section~\ref{sec:method_masking}. The segmentor remains frozen during pre-training and produces a vascular probability map for each input. The metric $\mathcal{L}$ compares the maps extracted from $I$ and $I^\prime$, and we use cross-entropy (CE) by default. We use the segmentor rather than Frangi filter in this loss because its thresholding and region-growing operations are non-differentiable and therefore prevent gradients from propagating to the model.

\textbf{Training Objective.} In addition to the anatomical consistency loss, we also use the pixel-level reconstruction loss, \textit{i.e.}, MSE, following conventional MIM approaches~\cite{he2022masked,kang2024deblurring,wang2025bootstrap}. The overall training objective is:
\begin{align}
    \mathcal{L}_{\rm MIM} = \mathcal{L}_{\rm rec.}+\mathcal{L}_{\rm cons.}
\end{align}

\section{Results} \label{sec:results}
\subsection{Datasets}
\begin{figure}[htbp]
\centering\centerline{\includegraphics{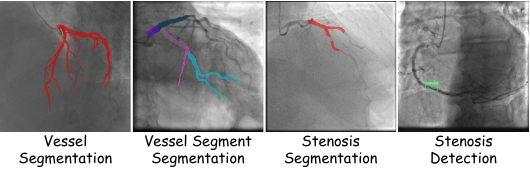}}
\caption{Overview of downstream tasks. We evaluate VasoMIM on four distinct tasks: vessel segmentation, vessel segment segmentation, stenosis segmentation, and stenosis detection.}
\label{fig:downstream}
\end{figure}

Table~\ref{table:dataset} summarizes the datasets used in the experiments, with downstream tasks illustrated in Fig.~\ref{fig:downstream}.

\textbf{Pre-training.} XA-$170$K is collected from four publicly available sources: CADICA~\cite{jimenez2024cadica}, SYNTAX~\cite{mahmoudi2025x}, XCAD~\cite{ma2021self}, and CoronaryDominance~\cite{kruzhilov2025coronarydominance}. \textbf{\textit{i)} CADICA} comprises coronary angiography videos from $42$ patients, with lengths ranging from $1$ to $151$ frames. From these, we select $6,594$ high-quality frames. \textbf{\textit{ii)} SYNTAX} contains $2,943$ X-ray angiograms derived from $231$ patients. \textbf{\textit{iii)} XCAD} provides a set of $1,747$ angiograms, from which $1,621$ images are utilized. \textbf{\textit{iv)} CoronaryDominance} consists of videos from $1,574$ patients. We extract informative frames from each video sequence, yielding a total of $160,320$ images.

\textbf{Segmentation.} We evaluate VasoMIM across three distinct segmentation tasks: vessel segmentation (ARCADE-V, CAXF, and XACV), vessel segment segmentation (ARCADE-VS), and stenosis segmentation (ARCADE-S). \textbf{\textit{i)} ARCADE} is divided into two subsets: one dedicated to vessel (V) and vessel segment (VS) segmentation, and another for stenosis (S) segmentation. Each subset contains $1,300$ images, partitioned into training ($1,000$) and testing ($300$) sets. \textbf{\textit{ii)} CAXF} consists of $538$ images derived from $36$ angiography videos. Following~\cite{li2020cau,huang2026spironet}, $337$ frames from $24$ videos are utilized for training, while the remaining $201$ images from $12$ videos are reserved for testing. Notably, the split is performed at the patient level, ensuring that no patient appears in both the training and testing sets. \textbf{\textit{iii)} XACV} comprises $111$ videos from $59$ patients. Frames with pixel-level vessel annotations are extracted and split by patient ID, resulting in $175$ training frames ($47$ patients) and $46$ testing frames ($12$ patients).

\textbf{Detection.} For this task, we utilize \textbf{Stenosis}, which contains angiography videos from $100$ patients totaling $8,325$ frames. The dataset is split into a training set of $7,492$ frames ($90$ patients) and a testing set of $833$ frames ($10$ patients).

\begin{table*}[t]
\caption{Quantitative comparison on segmentation tasks. We report ``${\rm mean}\pm{\rm std}$'' over five random seeds for all learning-based methods. Since Frangi filter is deterministic, we report only ``${\rm mean}$''. The best results are highlighted in \textbf{bold}, and the second-best are \underline{underlined}. Paired $t$-tests are conducted between VasoMIM-v1 and VasoMIM based on five seed-level results. $\ast$: $p<0.05$, $\ast\ast$: $p<0.01$, $\ast\ast\ast$: $p<0.001$.}
\label{table:segmentation_sota_compare}
\centering

\resizebox{\linewidth}{!}{
\begin{tabular}{llllllll|l|l|l}
\toprule
\multirow{2}{*}{Method} & \multirow{2}{*}{Pre-training Data} & \multicolumn{2}{c}{ARCADE-V} & \multicolumn{2}{c}{CAXF} & \multicolumn{2}{c}{XACV} & \multicolumn{1}{|c}{ARCADE-S} & \multicolumn{1}{|c|}{ARCADE-VS} & \multirow{2}{*}{Avg. Rank} \\ \cmidrule{3-10}
& & DSC (\%) & clDice (\%) & DSC (\%) & clDice (\%) & DSC (\%) & clDice (\%) & DSC (\%) & DSC (\%) \\ \midrule
\textit{Traditional} & & & & & & & & & \\
Frangi Filter~\cite{frangi1998multiscale} {\tiny\color{gray}[\textit{MICCAI}'98]} & $-$ & $41.30$ & $40.91$ & $64.01$ & $65.73$ & $58.46$ & $57.15$ & $-$ & $-$ & $-$ \\ \midrule
\textit{From Scratch} & & & & & & & & & \\
UNet~\cite{ronneberger2015u} {\tiny\color{gray}[\textit{MICCAI}'15]} & $-$ & $71.44${\tiny$\pm0.32$} & $70.67${\tiny$\pm0.40$} & $82.76${\tiny$\pm0.75$} & $84.95${\tiny$\pm0.56$} & $78.18${\tiny$\pm0.34$} & $74.71${\tiny$\pm0.44$} & $27.04${\tiny$\pm3.65$} & $38.77${\tiny$\pm0.82$} & $23.00${\tiny$\pm0.93$} \\ 
nnUNet~\cite{isensee2021nnu} {\tiny\color{gray}[\textit{Nat. Methods}'21]} & $-$ & $75.85${\tiny$\pm0.40$} & $77.82${\tiny$\pm0.16$} & $87.26${\tiny$\pm0.18$} & $90.31${\tiny$\pm0.26$} & $84.05${\tiny$\pm0.17$} & $82.63${\tiny$\pm0.15$} & $43.14${\tiny$\pm1.31$}& $47.86${\tiny$\pm2.37$} & $13.88${\tiny$\pm4.55$} \\
TransUNet~\cite{chen2024transunet} {\tiny\color{gray}[\textit{MedIA}'24]} & $-$ & $75.51${\tiny$\pm0.31$} & $76.29${\tiny$\pm0.37$} & $87.90${\tiny$\pm0.08$} & $90.98${\tiny$\pm0.15$} & $82.49${\tiny$\pm0.98$} & $81.00${\tiny$\pm1.01$} & $46.94${\tiny$\pm0.81$} & $50.46${\tiny$\pm1.29$} & $14.50${\tiny$\pm3.21$} \\
VM-UNet~\cite{ruan2025vm} {\tiny\color{gray}[\textit{ACM TOMM}'25]} & $-$ & $72.62${\tiny$\pm2.58$} & $72.82${\tiny$\pm3.19$} & $86.14${\tiny$\pm0.21$} & $90.11${\tiny$\pm0.20$} & $81.42${\tiny$\pm0.60$} & $79.02${\tiny$\pm0.72$} & $36.92${\tiny$\pm1.51$} & $44.39${\tiny$\pm2.78$} & $19.25${\tiny$\pm1.83$} \\
RWKV-UNet~\cite{jiang2025rwkv} {\tiny\color{gray}[\textit{arXiv}'25]} & $-$ & $70.56${\tiny$\pm0.88$} & $72.45${\tiny$\pm0.81$} & $83.96${\tiny$\pm1.02$} & $88.02${\tiny$\pm1.26$} & $81.14${\tiny$\pm0.57$} & $79.43${\tiny$\pm0.74$} & $32.79${\tiny$\pm4.19$} & $50.42${\tiny$\pm0.38$} & $20.38${\tiny$\pm2.00$} \\
\midrule
\textit{General SSL} & & & & & & & & & \\
MoCo v3~\cite{chen2021empirical} {\tiny\color{gray}[\textit{ICCV}'21]} & XA-$170$K & $76.08${\tiny$\pm0.19$} & $76.59${\tiny$\pm0.21$} & $87.20${\tiny$\pm0.36$} & $89.27${\tiny$\pm0.37$} & $83.75${\tiny$\pm0.16$} & $80.79${\tiny$\pm0.21$} & $41.71${\tiny$\pm0.81$} & $49.95${\tiny$\pm0.31$} & $16.00${\tiny$\pm2.45$} \\ 
DINO~\cite{caron2021emerging} {\tiny\color{gray}[\textit{ICCV}'21]} & XA-$170$K & $76.37${\tiny$\pm0.17$} & $77.04${\tiny$\pm0.26$} & $87.07${\tiny$\pm0.35$} & $89.32${\tiny$\pm0.41$} & $82.28${\tiny$\pm0.74$} & $79.50${\tiny$\pm0.88$} & $43.39${\tiny$\pm1.11$} & $51.08${\tiny$\pm0.43$} & $16.50${\tiny$\pm1.20$} \\ 
iBOT~\cite{zhou2022ibot} {\tiny\color{gray}[\textit{ICLR}'22]} & XA-$170$K & $76.58${\tiny$\pm0.26$} & $77.28${\tiny$\pm0.30$} & $86.50${\tiny$\pm0.40$} & $88.89${\tiny$\pm0.28$} & $80.84${\tiny$\pm0.92$} & $77.54${\tiny$\pm1.20$} & $44.50${\tiny$\pm0.61$} & $54.04${\tiny$\pm0.45$} & $17.00${\tiny$\pm3.16$} \\ 
MAE~\cite{he2022masked} {\tiny\color{gray}[\textit{CVPR}'22]} & XA-$170$K & $79.39${\tiny$\pm0.15$} & $80.74${\tiny$\pm0.20$} & $89.20${\tiny$\pm0.31$} & $91.80${\tiny$\pm0.36$} & $84.84${\tiny$\pm1.79$} & $82.58${\tiny$\pm1.95$} & $51.72${\tiny$\pm0.54$} & $56.69${\tiny$\pm0.48$} & $5.00${\tiny$\pm1.41$} \\ 
SimMIM~\cite{xie2022simmim} {\tiny\color{gray}[\textit{CVPR}'22]} & XA-$170$K & $77.81${\tiny$\pm0.22$} & $78.84${\tiny$\pm0.34$} & $88.02${\tiny$\pm0.63$} & $90.47${\tiny$\pm0.66$} & $82.77${\tiny$\pm0.79$} & $80.22${\tiny$\pm0.91$} & $47.63${\tiny$\pm0.69$} & $55.17${\tiny$\pm0.26$} & $12.75${\tiny$\pm2.25$} \\ 
AMT~\cite{liu2023good} {\tiny\color{gray}[\textit{AAAI}'23]} & XA-$170$K & $78.39${\tiny$\pm0.17$} & $79.58${\tiny$\pm0.16$} & $89.15${\tiny$\pm0.12$} & $91.77${\tiny$\pm0.17$} & $84.98${\tiny$\pm0.11$} & $82.58${\tiny$\pm0.16$} & $48.49${\tiny$\pm0.88$} & $55.68${\tiny$\pm0.39$} & $7.50${\tiny$\pm2.14$} \\ 
LocalMIM~\cite{wang2023masked} {\tiny\color{gray}[\textit{CVPR}'23]} & XA-$170$K & $78.79${\tiny$\pm0.20$} & $80.24${\tiny$\pm0.24$} & $88.44${\tiny$\pm0.70$} & $90.87${\tiny$\pm0.89$} & $83.18${\tiny$\pm1.84$} & $80.60${\tiny$\pm2.10$} & $51.20${\tiny$\pm0.45$} & $57.05${\tiny$\pm2.60$} & $9.50${\tiny$\pm3.96$} \\ 
I-JEPA~\cite{assran2023self} {\tiny\color{gray}[\textit{CVPR}'23]} & XA-$170$K & $77.06${\tiny$\pm0.17$} & $77.80${\tiny$\pm0.25$} & $88.05${\tiny$\pm0.12$} & $90.09${\tiny$\pm0.15$} & $84.30${\tiny$\pm0.16$} & $81.23${\tiny$\pm0.11$} & $47.38${\tiny$\pm0.86$} & $51.87${\tiny$\pm0.39$} & $12.50${\tiny$\pm2.51$} \\ 
HPM~\cite{wang2025bootstrap} {\tiny\color{gray}[\textit{TPAMI}'25]} & XA-$170$K & $78.84${\tiny$\pm0.15$} & $80.25${\tiny$\pm0.14$} & $88.86${\tiny$\pm0.24$} & $91.48${\tiny$\pm0.41$} & $85.19${\tiny$\pm0.07$} & $83.13${\tiny$\pm0.13$} & $50.06${\tiny$\pm0.29$} & $56.81${\tiny$\pm0.37$} & $6.12${\tiny$\pm1.81$} \\ 
DINOv3~\cite{simeoni2025dinov3} {\tiny\color{gray}[\textit{arXiv}'25]} & LVD-$1698$M & $79.36${\tiny$\pm0.07$} & $80.90${\tiny$\pm0.23$} & $89.13${\tiny$\pm0.26$} & $\underline{92.27}${\tiny$\pm0.44$} & $82.76${\tiny$\pm2.28$} & $80.62${\tiny$\pm2.40$} & $53.57${\tiny$\pm0.81$} & $54.36${\tiny$\pm0.59$} & $7.50${\tiny$\pm5.58$} \\ 
\midrule
\textit{Medical SSL} & & & & & & & & & \\
Models Genesis$^\dagger$~\cite{zhou2021models} {\tiny\color{gray}[\textit{MedIA}'21]} & XA-$170$K & $71.11${\tiny$\pm0.21$} & $70.75${\tiny$\pm0.19$} & $83.74${\tiny$\pm0.36$} & $85.49${\tiny$\pm0.41$} & $79.74${\tiny$\pm0.27$} & $76.34${\tiny$\pm0.24$} & $23.51${\tiny$\pm0.66$} & $41.26${\tiny$\pm0.46$} & $22.25${\tiny$\pm0.71$} \\ 
LVM-Med~\cite{mh2023lvm} {\tiny\color{gray}[\textit{NeurIPS}'23]} & Medical-$1.3$M & $74.82${\tiny$\pm0.20$} & $75.57${\tiny$\pm0.27$} & $85.21${\tiny$\pm0.98$} & $87.63${\tiny$\pm1.23$} & $82.47${\tiny$\pm0.81$} & $80.82${\tiny$\pm0.80$} & $39.34${\tiny$\pm5.10$} & $48.09${\tiny$\pm2.01$} & $18.38${\tiny$\pm2.45$} \\ 
DeblurringMIM~\cite{kang2024deblurring} {\tiny\color{gray}[\textit{MedIA}'24]} & XA-$170$K & $79.25${\tiny$\pm0.13$} & $80.77${\tiny$\pm0.12$} & $89.28${\tiny$\pm0.26$} & $91.91${\tiny$\pm0.35$} & $85.38${\tiny$\pm0.45$} & $83.10${\tiny$\pm0.59$} & $51.70${\tiny$\pm0.81$} & $56.66${\tiny$\pm0.52$} & $4.38${\tiny$\pm1.06$} \\ 
RAD-DINO~\cite{perez2025exploring} {\tiny\color{gray}[\textit{Nat. Mach. Intell.}'25]} & LVD-$142$M \& CXR-$838$K & $78.96${\tiny$\pm0.25$} & $80.26${\tiny$\pm0.40$} & $88.76${\tiny$\pm0.33$} & $91.66${\tiny$\pm0.59$} & $84.88${\tiny$\pm0.29$} & $83.29${\tiny$\pm0.47$} & $51.55${\tiny$\pm0.63$} & $54.81${\tiny$\pm0.15$} & $6.62${\tiny$\pm2.26$} \\
Frepa~\cite{chu2025improving} {\tiny\color{gray}[\textit{TMI}'25]} & XA-$170$K & $70.97${\tiny$\pm0.29$} & $70.07${\tiny$\pm0.54$} & $82.88${\tiny$\pm1.12$} & $84.79${\tiny$\pm1.24$} & $78.28${\tiny$\pm0.43$} & $74.67${\tiny$\pm0.60$} & $27.97${\tiny$\pm3.59$} & $38.12${\tiny$\pm1.16$} & $23.38${\tiny$\pm0.74$} \\
CheXWorld~\cite{yue2025chexworld} {\tiny\color{gray}[\textit{CVPR}'25]} & XA-$170$K & $78.18${\tiny$\pm0.24$} & $79.40${\tiny$\pm0.35$} & $88.52${\tiny$\pm0.22$} & $90.87${\tiny$\pm0.35$} & $83.81${\tiny$\pm0.79$} & $80.95${\tiny$\pm0.91$} & $49.69${\tiny$\pm0.65$} & $55.81${\tiny$\pm0.33$} & $10.12${\tiny$\pm1.55$} \\ 
MedDINOv3~\cite{li2025meddinov3} {\tiny\color{gray}[\textit{arXiv}'25]} & LVD-$1698$M \& CT-$3$M & $78.38${\tiny$\pm0.20$} & $79.53${\tiny$\pm0.29$} & $88.43${\tiny$\pm0.28$} & $90.96${\tiny$\pm0.38$} & $83.62${\tiny$\pm2.43$} & $81.24${\tiny$\pm2.86$} & $49.97${\tiny$\pm0.52$} & $54.98${\tiny$\pm0.38$} & $10.12${\tiny$\pm0.99$} \\ \midrule
\rowcolor{lightblue}VasoMIM-v1~\cite{huang2026vasomim} {\tiny\color{gray}[\textit{AAAI}'26]} & XA-$170$K & $\underline{79.90}${\tiny$\pm0.16$}$^{\ast\ast}$ & $\underline{81.57}${\tiny$\pm0.28$}$^{\ast\ast}$ & $\underline{89.36}${\tiny$\pm0.20$}$^{\ast}$ & $92.19${\tiny$\pm0.36$}$^{\ast}$ & $\underline{85.80}${\tiny$\pm0.27$}$^{\ast}$ & $\underline{83.75}${\tiny$\pm0.37$}$^{\ast}$ & $\underline{54.52}${\tiny$\pm0.53$}$^{\ast}$ & $\underline{58.03}${\tiny$\pm0.27$}$^{\ast\ast\ast}$ & $\underline{2.12}${\tiny$\pm0.35$} \\ 
\rowcolor{lightblue}VasoMIM {\tiny\textbf{\color{gray}[\textit{Ours}]}} & XA-$170$K & $\bm{80.25}${\tiny$\pm0.12$} & $\bm{82.06}${\tiny$\pm0.18$} & $\bm{89.68}${\tiny$\pm0.08$} & $\bm{92.57}${\tiny$\pm0.12$} & $\bm{86.09}${\tiny$\pm0.09$} & $\bm{84.12}${\tiny$\pm0.17$} & $\bm{55.62}${\tiny$\pm0.63$} & $\bm{58.87}${\tiny$\pm0.15$} & $\bm{1.00}${\tiny$\pm0.00$} \\ 
\rowcolor{lightblue}$\Delta$ \textit{vs.} UNet & $-$ & {\color{deepgreen}$\uparrow\bm{8.81}$} & {\color{deepgreen}$\uparrow\bm{11.39}$} & {\color{deepgreen}$\uparrow\bm{6.92}$} & {\color{deepgreen}$\uparrow\bm{7.62}$} & {\color{deepgreen}$\uparrow\bm{7.91}$} & {\color{deepgreen}$\uparrow\bm{9.41}$} & {\color{deepgreen}$\uparrow\bm{28.58}$} & {\color{deepgreen}$\uparrow\bm{20.10}$} & $-$ \\
\bottomrule
\multicolumn{10}{l}{$\dagger$: We implement the 2D version of Models Genesis based on the official codebase.}
\end{tabular}
}

\end{table*}

\subsection{Implementation Details}
\textbf{Pre-training.} The pre-training configuration largely adheres to the standard MAE protocol~\cite{he2022masked}. ViT-B/$16$~\cite{dosovitskiy2021an} is utilized as the default backbone unless stated otherwise. We adopt the official implementation of UNeXt-S~\cite{valanarasu2022unext} to construct the segmentor, which is trained on XA-$170$K using pseudo-labels generated by Frangi filter and remains frozen throughout the pre-training process. Optimization is performed using AdamW~\cite{loshchilov2019decoupled} with a base learning rate ${\rm lr}_{\rm base}$ of $1.5e^{-4}$. Following~\cite{he2022masked}, the linear learning rate scaling rule is applied: ${\rm lr} = {\rm lr}_{\rm base} \times \text{batch\_size} / 256$. VasoMIM is pre-trained for $800$ epochs with an input resolution of $224 \times 224$ and a batch size of $256$ per GPU. This process requires approximately ten hours on $8\times$NVIDIA H$20$ GPUs.

\textbf{Segmentation.} We adopt UNet~\cite{ronneberger2015u} as the segmentation framework, performing end-to-end fine-tuning at a resolution of $224\times224$. To enhance feature aggregation, we upgrade the standard UNet decoder used in our previous work~\cite{huang2026vasomim} by incorporating a SimpleFPN~\cite{li2022exploring} adapter, following the implementation in CheXWorld~\cite{yue2025chexworld}. Optimization is performed using AdamW with a cosine-annealing schedule, where $T_{\rm max}={\rm Epoch}_{\rm max}$. All downstream segmentation experiments are conducted on a single NVIDIA H$20$ GPU. The specific hyper-parameters for each dataset are detailed below:
\begin{itemize}
    \item ARCADE-V: ${\rm lr}_{\rm base}=2e^{-4}$, ${\rm Epoch}_{\rm max}=100$
    \item CAXF: ${\rm lr}_{\rm base}=4e^{-4}$, ${\rm Epoch}_{\rm max}=200$
    \item XACV: ${\rm lr}_{\rm base}=6e^{-4}$, ${\rm Epoch}_{\rm max}=200$
    \item ARCADE-S: ${\rm lr}_{\rm base}=8e^{-5}$, ${\rm Epoch}_{\rm max}=100$
    \item ARCADE-VS: ${\rm lr}_{\rm base}=1e^{-4}$, ${\rm Epoch}_{\rm max}=100$
\end{itemize}

\textbf{Detection.} For stenosis detection, Faster R-CNN~\cite{ren2016faster} equipped with a feature pyramid network (FPN)~\cite{lin2017feature} is employed. To generate the requisite pyramid feature maps, the transformer blocks of ViT-B/$16$ are evenly divided into four subsets, following the protocols established in~\cite{he2022masked,wang2025bootstrap}. Convolutional operations convert the output of each group into a pyramid feature map with a stride of $4$, $8$, $16$, or $32$. The architecture is fine-tuned end-to-end for $100$ epochs at a resolution of $512\times512$ utilizing a single NVIDIA H$20$ GPU. All settings strictly adhere to the standard ViTDet~\cite{li2022exploring} configurations implemented within Detectron2\footnote{\href{https://github.com/facebookresearch/detectron2}{https://github.com/facebookresearch/detectron2}}.

\textbf{Evaluation Metrics.} For all segmentation tasks, we report the Dice similarity coefficient (DSC). Additionally, given the thin and curvilinear characteristics of vascular structures, we report centerlineDice (clDice)~\cite{shit2021cldice} exclusively for vessel segmentation tasks to better quantify topological correctness. For detection tasks, we report the mean average precision (mAP), including mAP${50}$ and mAP${75}$, as well as the mAP averaged across IoU thresholds from $0.50$ to $0.95$.

\subsection{Main Results}
All baselines are implemented using their official codebases. To ensure a rigorous comparison, each model is fine-tuned under identical settings across five random seeds. We report all metrics as ``${\rm mean} \pm {\rm std}$''. Moreover, all SSL methods use the same backbone architecture, \textit{i.e.}, ViT-B/$16$.

\begin{figure*}[htbp]
    \centering
    \begin{subfigure}[c]{0.45\textwidth} 
        \centering
        \includegraphics{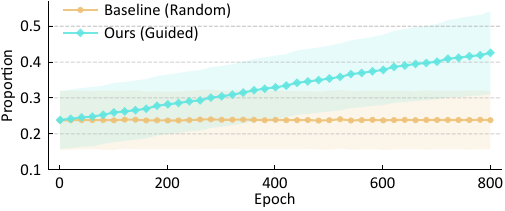}
    \end{subfigure}
    \hspace{0.2cm}
    \begin{subfigure}[c]{0.52\textwidth}
        \centering
        \includegraphics{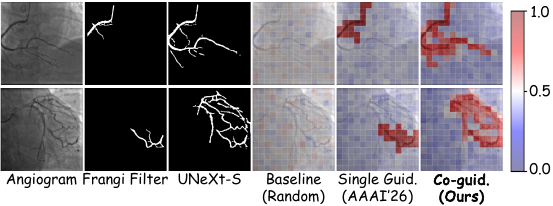}
    \end{subfigure}
    \caption{Qualitative analysis of the anatomy-guided masking strategy. \textbf{Left:} The proportion of vessel-containing patches among all masked patches across training epochs. \textbf{Right:} Visual comparison. We highlight failure cases of Frangi filter where UNeXt-S~\cite{valanarasu2022unext}, leveraging a strong inductive bias, extracts vascular anatomy more accurately to provide precise guidance. The heatmaps display the patch-wise masking ratio, \textit{i.e.}, $\frac{1}{E}\sum_{j=1}^E\mathbb{I}\left(\text{Patch $x_i$ is masked in epoch $j$}\right)$.}
    \label{fig:segguidance}
\end{figure*}
\textbf{Segmentation.} Several key findings can be observed from Table~\ref{table:segmentation_sota_compare}: \textbf{\textit{i)} Pre-training unlocks the potential of segmentors.} By pre-training on large-scale unlabeled data, a standard UNet~\cite{ronneberger2015u} can outperform advanced segmentors with complex architectures trained from scratch~\cite{isensee2021nnu,chen2024transunet,ruan2025vm,jiang2025rwkv}. \textbf{\textit{ii)} VasoMIM establishes a new state-of-the-art.} Our method consistently surpasses all general and medical SSL alternatives across every downstream dataset. Notably, even DINOv3~\cite{simeoni2025dinov3}, the leading vision foundation model pre-trained on nearly $1.7$ billion images, lags behind VasoMIM, \textit{e.g.}, $79.36\%$ \textit{vs}. $80.25\%$ DSC on ARCADE-V. This underscores the critical necessity of domain-specific pre-training. \textbf{\textit{iii)} Robustness on challenging tasks.} The advantage of VasoMIM is most pronounced on complex, fine-grained tasks such as stenosis segmentation (ARCADE-S) and vessel segment segmentation (ARCADE-VS). This widening performance gap highlights VasoMIM’s superior generalizability and its ability to reliably capture fine-grained topological details where baselines are less effective.

\begin{table}[htbp]
\caption{Quantitative results for the stenosis detection task. We report ``${\rm mean}\pm{\rm std}$'' over five random seeds. The best and second-best results are highlighted in \textbf{bold} and \underline{underlined}, respectively. Paired $t$-tests are conducted between VasoMIM-v1 and VasoMIM based on five seed-level results. $\ast$: $p<0.05$, $\ast\ast$: $p<0.01$, ${\rm n.s.}$: not significant.}
\label{table:detection_sota_compare}
\centering
\resizebox{\linewidth}{!}{
\begin{tabular}{llllll}
\toprule
Method & Pre-training Data & mAP$50$ (\%) & mAP$75$ (\%) & mAP (\%) & Avg. Rank \\ \midrule
\multicolumn{3}{l}{\textit{From Scratch}} \\ 
Faster R-CNN~\cite{ren2016faster} {\tiny\color{gray}[\textit{TPAMI}'16]} & $-$ & $88.37${\tiny$\pm0.90$} & $19.01${\tiny$\pm1.77$} & $36.63${\tiny$\pm0.63$} & $20.33${\tiny$\pm0.58$} \\
DETR~\cite{carion2020end} {\tiny\color{gray}[\textit{ECCV}'20]} & $-$ & $89.30${\tiny$\pm0.79$} & $19.35${\tiny$\pm2.00$} & $36.51${\tiny$\pm0.57$} & $19.00${\tiny$\pm2.65$} \\
\midrule
\multicolumn{3}{l}{\textit{General SSL}}  \\
MoCo v3~\cite{chen2021empirical} {\tiny\color{gray}[\textit{ICCV}'21]} & XA-$170$K & $91.54${\tiny$\pm1.04$} & $21.74${\tiny$\pm0.67$} & $38.74${\tiny$\pm0.32$} & $14.00${\tiny$\pm1.00$} \\ 
DINO~\cite{caron2021emerging} {\tiny\color{gray}[\textit{ICCV}'21]} & XA-$170$K & $88.83${\tiny$\pm0.46$} & $21.12${\tiny$\pm0.78$} & $37.65${\tiny$\pm0.41$} & $17.00${\tiny$\pm1.00$} \\ 
iBOT~\cite{zhou2022ibot} {\tiny\color{gray}[\textit{ICLR}'22]} & XA-$170$K & $92.16${\tiny$\pm0.85$} & $23.94${\tiny$\pm0.68$} & $39.76${\tiny$\pm0.30$} & $9.67${\tiny$\pm2.08$} \\
MAE~\cite{he2022masked} {\tiny\color{gray}[\textit{CVPR}'22]} & XA-$170$K & $92.30${\tiny$\pm0.42$} & $24.28${\tiny$\pm0.60$} & $39.69${\tiny$\pm0.48$} & $9.67${\tiny$\pm2.31$} \\
SimMIM~\cite{xie2022simmim} {\tiny\color{gray}[\textit{CVPR}'22]} & XA-$170$K & $89.84${\tiny$\pm0.72$} & $22.30${\tiny$\pm0.27$} & $38.29${\tiny$\pm0.38$} & $15.00${\tiny$\pm1.00$} \\ 
AMT~\cite{liu2023good} {\tiny\color{gray}[\textit{AAAI}'23]} & XA-$170$K & $93.35${\tiny$\pm1.20$} & $24.33${\tiny$\pm2.08$} & $40.23${\tiny$\pm0.59$} & $7.00${\tiny$\pm1.00$} \\
LocalMIM~\cite{wang2023masked} {\tiny\color{gray}[\textit{CVPR}'23]} & XA-$170$K & $93.59${\tiny$\pm0.53$} & $24.76${\tiny$\pm0.47$} & $40.24${\tiny$\pm0.61$} & $5.00${\tiny$\pm1.73$} \\
I-JEPA~\cite{assran2023self} {\tiny\color{gray}[\textit{CVPR}'23]} & XA-$170$K & $90.47${\tiny$\pm0.89$} & $22.51${\tiny$\pm1.36$} & $38.63${\tiny$\pm0.75$} & $14.00${\tiny$\pm1.00$} \\
HPM~\cite{wang2025bootstrap} {\tiny\color{gray}[\textit{TPAMI}'25]} & XA-$170$K & $92.47${\tiny$\pm0.36$} & $24.65${\tiny$\pm1.12$} & $40.53${\tiny$\pm0.35$} & $6.67${\tiny$\pm2.89$} \\
DINOv3~\cite{simeoni2025dinov3} {\tiny\color{gray}[\textit{arXiv}'25]} & LVD-$1698$M & $93.89${\tiny$\pm0.81$} & $23.60${\tiny$\pm1.87$} & $40.90${\tiny$\pm0.39$} & $5.33${\tiny$\pm4.04$} \\
\midrule
\multicolumn{3}{l}{\textit{Medical SSL}}  \\
Models Genesis$^\dagger$~\cite{zhou2021models} {\tiny\color{gray}[\textit{MedIA}'21]} & XA-$170$K & $87.81${\tiny$\pm1.06$} & $19.96${\tiny$\pm0.81$} & $36.89${\tiny$\pm0.46$} & $19.33${\tiny$\pm1.53$} \\ 
LVM-Med~\cite{mh2023lvm} {\tiny\color{gray}[\textit{NeurIPS}'23]} & Medical-$1.3$M & $93.64${\tiny$\pm0.88$} & $23.14${\tiny$\pm0.89$} & $39.75${\tiny$\pm0.40$} & $9.00${\tiny$\pm3.61$} \\
DeblurringMIM~\cite{kang2024deblurring} {\tiny\color{gray}[\textit{MedIA}'24]} & XA-$170$K & $88.99${\tiny$\pm0.96$} & $24.73${\tiny$\pm1.60$} & $39.26${\tiny$\pm0.65$} & $11.33${\tiny$\pm6.66$} \\
RAD-DINO~\cite{perez2025exploring} {\tiny\color{gray}[\textit{Nat. Mach. Intell.}'25]} & LVD-$142$M \& CXR-$838$K & $93.56${\tiny$\pm0.25$} & $23.92${\tiny$\pm1.33$} & $40.64${\tiny$\pm0.63$} & $6.67${\tiny$\pm2.52$} \\
Frepa~\cite{chu2025improving} {\tiny\color{gray}[\textit{TMI}'25]} & XA-$170$K & $88.61${\tiny$\pm0.82$} & $19.82${\tiny$\pm0.88$} & $37.22${\tiny$\pm0.27$} & $18.67${\tiny$\pm0.58$} \\
CheXWorld~\cite{yue2025chexworld} {\tiny\color{gray}[\textit{CVPR}'25]} & XA-$170$K & $92.89${\tiny$\pm1.02$} & $23.48${\tiny$\pm1.11$} & $40.16${\tiny$\pm0.98$} & $9.33${\tiny$\pm1.53$} \\
MedDINOv3~\cite{li2025meddinov3} {\tiny\color{gray}[\textit{arXiv}'25]} & LVD-$1698$M \& CT-$3$M & $93.79${\tiny$\pm0.47$} & $20.96${\tiny$\pm1.72$} & $39.69${\tiny$\pm0.64$} & $10.67${\tiny$\pm6.51$} \\ \midrule
\rowcolor{lightblue}VasoMIM-v1~\cite{huang2026vasomim} {\tiny\color{gray}[\textit{AAAI}'26]} & XA-$170$K & $\underline{94.25}${\tiny$\pm0.72$}$^{\ast}$ & $\underline{25.01}${\tiny$\pm1.25$}$^{\ast\ast}$ & $\underline{40.91}${\tiny$\pm0.63$}$^{\rm n.s.}$ & $\underline{2.00}${\tiny$\pm0.00$} \\
\rowcolor{lightblue}VasoMIM {\tiny\textbf{\color{gray}[\textit{Ours}]}} & XA-$170$K & $\bm{94.91}${\tiny$\pm0.28$} & $\bm{25.72}${\tiny$\pm1.57$} & $\bm{41.07}${\tiny$\pm0.35$} & $\bm{1.00}${\tiny$\pm0.00$} \\
\rowcolor{lightblue}$\Delta$ \textit{vs.} Faster R-CNN & $-$ & {\color{deepgreen}$\uparrow\bm{6.54}$} & {\color{deepgreen}$\uparrow\bm{6.71}$} & {\color{deepgreen}$\uparrow\bm{4.44}$} & $-$ \\
\bottomrule
\multicolumn{6}{l}{$\dagger$: We implement the 2D version of Models Genesis based on the official codebase.}
\end{tabular}
}
\end{table}

\textbf{Detection.} Table~\ref{table:detection_sota_compare} presents the quantitative results for the stenosis detection task. VasoMIM delivers substantial improvements over the supervised baseline, yielding gains of $+6.54\%$ mAP$50$, $+6.71\%$ mAP$75$, and $+4.44\%$ mAP compared to Faster R-CNN~\cite{ren2016faster} trained from scratch. Among the SSL baselines, LocalMIM~\cite{wang2023masked} emerges as a strong competitor. Nevertheless, VasoMIM surpasses this method by clear margins, achieving further gains of $+1.32\%$ mAP$50$, $+0.96\%$ mAP$75$, and $+0.83\%$ mAP. These results collectively underscore the superiority of the proposed domain-specific pre-training framework in learning robust, transferable representations for localizing complex lesion features.

\textbf{Overall Improvements over VasoMIM-v1~\cite{huang2026vasomim}.} Across all downstream tasks, VasoMIM consistently outperforms VasoMIM-v1. To evaluate the statistical significance of these improvements, we perform paired $t$-tests over five matched random seeds for each dataset and evaluation metric reported in Tables~\ref{table:segmentation_sota_compare} and~\ref{table:detection_sota_compare}. Statistically significant improvements ($p<0.05$) are observed in $10$ out of the $11$ comparisons. These results demonstrate that the performance gains of VasoMIM over VasoMIM-v1 are statistically reliable across the majority of downstream evaluations.

\begin{table}[htbp]
\caption{Ablation study of the anatomy-guided masking strategy and the anatomical consistency loss. Paired $t$-tests are conducted between each variant and VasoMIM using five seed-level results. $\ast$: $p<0.05$, $\ast\ast$: $p<0.01$, $\ast\ast\ast$: $p<0.001$, ${\rm n.s.}$: not significant.}
\label{table:ablation}
\centering
\resizebox{\linewidth}{!}{
\begin{tabular}{llllll}
\toprule
\multirow{2}{*}{Guidance} & \multirow{2}{*}{$\mathcal{L}_{\rm cons.}$} & \multicolumn{2}{c}{ARCADE-V} & \multicolumn{2}{c}{XACV} \\ \cmidrule{3-6} 
& & DSC (\%) & clDice (\%) & DSC (\%) & clDice (\%) \\ \midrule
 $-$ & $-$ & $79.31${\tiny$\pm0.16$}$^{\ast\ast\ast}$ & $81.05${\tiny$\pm0.25$}$^{\ast\ast\ast}$ & $84.52${\tiny$\pm2.12$}$^{\rm n.s.}$ & $82.25${\tiny$\pm2.32$}$^{\rm n.s.}$ \\
\multirow{2}{*}{$-$} & \multirow{2}{*}{\checkmark} & $79.85${\tiny$\pm0.14$}$^{\ast}$ & $\underline{81.59}${\tiny$\pm0.16$}$^{\ast}$ & $85.79${\tiny$\pm0.52$}$^{\rm n.s.}$ & $83.66${\tiny$\pm0.77$}$^{\rm n.s.}$ \\
 & & {\color{deepgreen}$\uparrow0.54$} & {\color{deepgreen}$\uparrow0.54$} & {\color{deepgreen}$\uparrow1.27$} & {\color{deepgreen}$\uparrow1.41$} \\
\multirow{2}{*}{\checkmark} & \multirow{2}{*}{$-$} & $\underline{79.87}${\tiny$\pm0.09$}$^{\ast\ast}$ & $81.41${\tiny$\pm0.11$}$^{\ast\ast}$ & $\underline{85.92}${\tiny$\pm0.04$}$^{\ast}$ & $\underline{83.78}${\tiny$\pm0.15$}$^{\ast}$ \\
 & & {\color{deepgreen}$\uparrow0.56$} & {\color{deepgreen}$\uparrow0.36$} & {\color{deepgreen}$\uparrow1.40$} & {\color{deepgreen}$\uparrow1.53$} \\
 \cellcolor{lightblue} & \cellcolor{lightblue} & \cellcolor{lightblue}$\bm{80.25}${\tiny$\pm0.12$} & \cellcolor{lightblue}$\bm{82.06}${\tiny$\pm0.18$} & \cellcolor{lightblue}$\bm{86.09}${\tiny$\pm0.09$} & \cellcolor{lightblue}$\bm{84.12}${\tiny$\pm0.17$} \\
 \multirow{-2}{*}{\cellcolor{lightblue}\checkmark} & \multirow{-2}{*}{\cellcolor{lightblue}\checkmark} & \cellcolor{lightblue}{\color{deepgreen}$\uparrow\bm{0.94}$} & \cellcolor{lightblue}{\color{deepgreen}$\uparrow\bm{1.01}$} & \cellcolor{lightblue}{\color{deepgreen}$\uparrow\bm{1.57}$} & \cellcolor{lightblue}{\color{deepgreen}$\uparrow\bm{1.87}$} \\
 \bottomrule
\end{tabular}
}
\end{table}


\subsection{Ablation Study}
Ablation studies are conducted on ARCADE-V and XACV. Default settings are highlighted in \sethlcolor{lightblue}\hl{blue}. Table~\ref{table:ablation} reports the quantitative results of sequentially integrating each proposed component into the vanilla baseline, \textit{i.e.}, MAE ($\gamma=0.5$).

\textbf{Effectiveness of Anatomy-guided Masking Strategy.} The integration of the proposed anatomy-guided masking strategy yields a notable performance boost, specifically achieving absolute gains of $+0.56\%$ and $+1.40\%$ DSC on ARCADE-V and XACV, respectively. A detailed analysis is provided to elucidate the mechanisms driving these improvements. As illustrated in Fig.~\ref{fig:segguidance} (Left), the proposed strategy dynamically increases the proportion of masked vessel-containing patches throughout the pre-training process, whereas the random baseline maintains a relatively low and static proportion. Furthermore, we visualize the patch-wise masking ratio in Fig.~\ref{fig:segguidance} (Right). Although the single-guidance strategy~\cite{huang2026vasomim} prioritizes vessel-containing patches, it frequently omits critical branches due to the inherent limitations of Frangi filter. In contrast, the introduced co-guidance strategy leverages the population-level vascular patterns learned by the segmentor to capture vascular anatomy with higher fidelity, ensuring the model focuses on the most discriminative and semantically rich regions.

\begin{table}[htbp]
\caption{Comparison with alternative reconstruction objectives. GFLOPs are calculated on an NVIDIA H$20$ GPU using a single $224\times224$ masked RGB image. DSC is reported in this table.}
\label{table:loss}
\centering
\resizebox{\linewidth}{!}{
    \begin{tabular}{lllllll}
    \toprule
    Loss & Params. (M) & FLOPs (G) & ARCADE-V & XACV \\ \midrule
    \multicolumn{5}{l}{\color{gray}\textit{Baseline}} \\
    {\color{gray}$-$} & {\color{gray}$111.91$} & {\color{gray}$19.01$} & {\color{gray}$79.87${\tiny$\pm0.09$}} & {\color{gray}$85.92${\tiny$\pm0.04$}} \\ \midrule
    \multicolumn{5}{l}{\textit{Feature Distillation}} \\
    \multirow{2}{*}{DINO~\cite{caron2021emerging} {\tiny\color{gray}[\textit{ICCV}'21]}} & $197.95$ & $52.90$ & $79.81${\tiny$\pm0.12$} & $85.03${\tiny$\pm2.27$} \\
    & {\color{red}$+86.04$} & {\color{red}$+33.89$} & {\color{red}$\downarrow 0.06$} & {\color{red}$\downarrow 0.89$} \\
    \multirow{2}{*}{CLIP~\cite{radford2021learning} {\tiny\color{gray}[\textit{ICML}'21]}} & $169.43$ & $41.66$ & $79.79${\tiny$\pm0.13$} & $85.29${\tiny$\pm1.47$} \\
    & {\color{red}$+57.52$} & {\color{red}$+22.65$} & {\color{red}$\downarrow 0.08$} & {\color{red}$\downarrow 0.63$} \\
    \multirow{2}{*}{DINOv3~\cite{simeoni2025dinov3} {\tiny\color{gray}[\textit{arXiv}'25]}} & $176.69$ & $45.04$ & $80.03${\tiny$\pm0.11$} & $\underline{86.04}${\tiny$\pm0.09$} \\
    & {\color{red}$+64.78$} & {\color{red}$+26.03$} & {\color{deepgreen}$\uparrow0.16$} & {\color{deepgreen}$\uparrow0.12$} \\
    \multirow{2}{*}{EMA} & $197.95$ & $52.90$ & $\underline{80.05}${\tiny$\pm0.12$} & $86.00${\tiny$\pm0.09$} \\
    & {\color{red}$+86.04$} & {\color{red}$+33.89$} & {\color{deepgreen}$\uparrow0.18$} & {\color{deepgreen}$\uparrow0.08$} \\ \midrule
    \multicolumn{5}{l}{\textit{Perceptual Loss}} \\
    \multirow{2}{*}{LPIPS~\cite{zhang2018unreasonable} {\tiny\color{gray}[\textit{CVPR}'18]}} & $126.63$ & $80.45$ & $79.18${\tiny$\pm0.14$} & $86.00${\tiny$\pm0.08$} \\
    & {\color{red}$+14.72$} & {\color{red}$+61.44$} & {\color{red}$\downarrow0.69$} & {\color{deepgreen}$\uparrow0.08$} \\
    \multirow{2}{*}{Contextual~\cite{mechrez2018contextual} {\tiny\color{gray}[\textit{ECCV}'18]}} & $131.93$ & $97.05$ & $79.90${\tiny$\pm0.10$} & $85.63${\tiny$\pm0.91$} \\
    & {\color{red}$+20.02$} & {\color{red}$+78.04$} & {\color{deepgreen}$\uparrow0.03$} & {\color{red}$\downarrow0.29$} \\
    \midrule
    \cellcolor{lightblue} & \cellcolor{lightblue}$111.91$ & \cellcolor{lightblue}$19.17$ & \cellcolor{lightblue}$\bm{80.25}${\tiny$\pm0.12$} & \cellcolor{lightblue}$\bm{86.09}${\tiny$\pm0.09$} \\
    \multirow{-2}{*}{\cellcolor{lightblue}$\mathcal{L}_{\rm cons.}$} & \cellcolor{lightblue}{\color{gray}$0.00$} & \cellcolor{lightblue}{\color{red}$+0.16$} & \cellcolor{lightblue}{\color{deepgreen}$\uparrow\bm{0.38}$} & \cellcolor{lightblue}{\color{deepgreen}$\uparrow\bm{0.17}$} \\ \bottomrule
    \end{tabular}
}
\end{table}
\begin{table}[htbp]
\caption{Comparison of clustering metrics on XACV \textit{training set}.}
\label{table:loss_cluster}
\centering
    \begin{tabular}{lllll}
    \toprule
    Loss & SS $\uparrow$ & CHI $\uparrow$ & DBI $\downarrow$ \\ \midrule
    \multicolumn{4}{l}{\color{gray}\textit{Baseline}} \\
    {\color{gray}$-$} & $0.27$ & $1.39$ & $1.37$  \\ \midrule
    \multicolumn{4}{l}{\textit{Feature Distillation}} \\
    DINO~\cite{caron2021emerging} {\tiny\color{gray}[\textit{ICCV}'21]} & $0.03$ & $0.06$ & $6.42$ \\
    CLIP~\cite{radford2021learning} {\tiny\color{gray}[\textit{ICML}'21]} & $0.08$ & $0.18$ & $3.71$ \\
    DINOv3~\cite{simeoni2025dinov3} {\tiny\color{gray}[\textit{arXiv}'25]} & $\underline{0.29}$ & $\underline{1.65}$ & $\underline{1.24}$ \\
    EMA & $0.27$ & $1.53$ & $1.33$ \\ \midrule
    \multicolumn{4}{l}{\textit{Perceptual Loss}} \\
    LPIPS~\cite{zhang2018unreasonable} {\tiny\color{gray}[\textit{CVPR}'18]} & $0.09$ & $0.18$ & $3.69$ \\ 
    Contextual~\cite{mechrez2018contextual} {\tiny\color{gray}[\textit{ECCV}'18]} & $0.26$ & $1.27$ & $1.42$ \\
    \midrule
    \rowcolor{lightblue}$\mathcal{L}_{\rm cons.}$ & $\bm{0.32}$ & $\bm{2.05}$ & $\bm{1.13}$ \\ \bottomrule
    \multicolumn{4}{l}{SS: Silhouette Score.} \\
    \multicolumn{4}{l}{CHI ($\times10^5$): Calinski-Harabasz Index.} \\
    \multicolumn{4}{l}{DBI: Davies-Bouldin Index.}
    \end{tabular}
\end{table}

\textbf{Significance of Anatomical Consistency Loss.} The anatomical consistency loss enhances the performance of the baseline by $+0.54\%$ and $+1.27\%$ DSC on ARCADE-V and XACV, respectively. To rigorously evaluate the superiority of $\mathcal{L}_{\rm cons.}$, we compare it against alternative reconstruction objectives in Table~\ref{table:loss}. Our method offers two distinct advantages: \textbf{\textit{i)} Superior Downstream Performance.} The proposed loss consistently outperforms both feature distillation and perceptual loss. \textbf{\textit{ii)}} \textbf{Negligible Computational Overhead.} Alternative objectives introduce substantial computational burdens. For instance, feature distillation via DINOv3 introduces $+64.78$ M parameters and $+26.03$ GFLOPs. In contrast, $\mathcal{L}_{\rm cons.}$ requires no additional parameters and incurs a trivial computational cost of merely $+0.16$ GFLOPs. Furthermore, the clustering properties of the learned representations are analyzed to assess feature discriminability. Image patches are first categorized into vessel-containing and background-only groups, followed by dimensionality reduction via t-SNE~\cite{maaten2008visualizing}. As reported in Table~\ref{table:loss_cluster}, the model trained with $\mathcal{L}_{\rm cons.}$ achieves the best clustering performance, indicating that it learns more discriminative vascular representations.

\begin{table}[htbp]
\caption{Comparison between instance-level and population-level anatomical guidance. Paired $t$-tests are conducted between each variant and VasoMIM using five seed-level results. $\ast$: $p<0.05$, $\ast\ast$: $p<0.01$, $\ast\ast\ast$: $p<0.001$, ${\rm n.s.}$: not significant.}~\label{table:guidance_comparison_aaai}
\centering
\resizebox{\linewidth}{!}{
\begin{tabular}{lllll}
\toprule
\multirow{2}{*}{Guidance} & \multicolumn{2}{c}{ARCADE-V} & \multicolumn{2}{c}{XACV} \\ \cmidrule{2-5} 
 & DSC (\%) & clDice (\%) & DSC (\%) & clDice (\%) \\ \midrule
$-$ & $79.85${\tiny$\pm0.14$}$^{\ast}$ & $81.59${\tiny$\pm0.16$}$^{\ast}$ & $85.79${\tiny$\pm0.52$}$^{\rm n.s.}$ & $83.66${\tiny$\pm0.77$}$^{\rm n.s.}$ \\
\multirow{2}{*}{Frangi-only~\cite{huang2026vasomim} {\tiny\color{gray}[\textit{AAAI}'26]}} & $79.90${\tiny$\pm0.16$}$^{\ast\ast}$ & $81.57${\tiny$\pm0.28$}$^{\ast\ast}$ & $85.80${\tiny$\pm0.27$}$^{\ast}$ & $83.75${\tiny$\pm0.37$}$^{\ast}$ \\
 & {\color{deepgreen}$\uparrow0.05$} & {\color{red}$\downarrow0.02$} & {\color{deepgreen}$\uparrow0.01$} & {\color{deepgreen}$\uparrow0.09$} \\
\multirow{2}{*}{Segmentor-only} & $\underline{79.98}${\tiny$\pm0.13$}$^{\ast\ast\ast}$ & $\underline{81.77}${\tiny$\pm0.19$}$^{\ast\ast}$ & $\underline{85.98}${\tiny$\pm0.08$}$^{\ast\ast}$ & $\underline{83.91}${\tiny$\pm0.13$}$^{\ast}$ \\
 & {\color{deepgreen}$\uparrow0.13$} & {\color{deepgreen}$\uparrow0.18$} & {\color{deepgreen}$\uparrow0.19$} & {\color{deepgreen}$\uparrow0.25$} \\
\cellcolor{lightblue} & \cellcolor{lightblue}$\bm{80.25}${\tiny$\pm0.12$} & \cellcolor{lightblue}$\bm{82.06}${\tiny$\pm0.18$} & \cellcolor{lightblue}$\bm{86.09}${\tiny$\pm0.09$} & \cellcolor{lightblue}$\bm{84.12}${\tiny$\pm0.17$} \\
\cellcolor{lightblue}\multirow{-2}{*}{Co-guidance} &\cellcolor{lightblue}{\color{deepgreen}$\uparrow\bm{0.40}$} & \cellcolor{lightblue}{\color{deepgreen}$\uparrow\bm{0.47}$} & \cellcolor{lightblue}{\color{deepgreen}$\uparrow\bm{0.30}$} & \cellcolor{lightblue}{\color{deepgreen}$\uparrow\bm{0.46}$} \\ \bottomrule
\end{tabular}}
\end{table}

\textbf{Complementary Effect of the Two Components.} Combining the two components achieves the best overall performance. We further conduct paired $t$-tests over five matched random seeds between each ablation variant and VasoMIM. VasoMIM significantly outperforms the guidance-only variant across all four metrics ($p<0.05$), and also significantly outperforms the $\mathcal{L}_{\rm cons.}$-only variant on ARCADE-V ($p<0.05$). On XACV, although the improvements over the $\mathcal{L}_{\rm cons.}$-only variant and the baseline do not reach statistical significance, both variants exhibit relatively large seed-level variation. The anatomical consistency loss reduces this variation to some extent, while incorporating anatomy-guided masking leads to a much more substantial improvement in seed stability. VasoMIM ultimately achieves the highest mean performance with consistently low variation.

\section{In-depth Analysis} \label{sec:analysis}
In this section, we present a comprehensive analysis of VasoMIM. Unless otherwise specified, all experiments are conducted on ARCADE-V and XACV. For clarity, the default configurations of VasoMIM are highlighted in \sethlcolor{lightblue}\hl{blue}.

\begin{figure*}[t]
\centering\centerline{\includegraphics{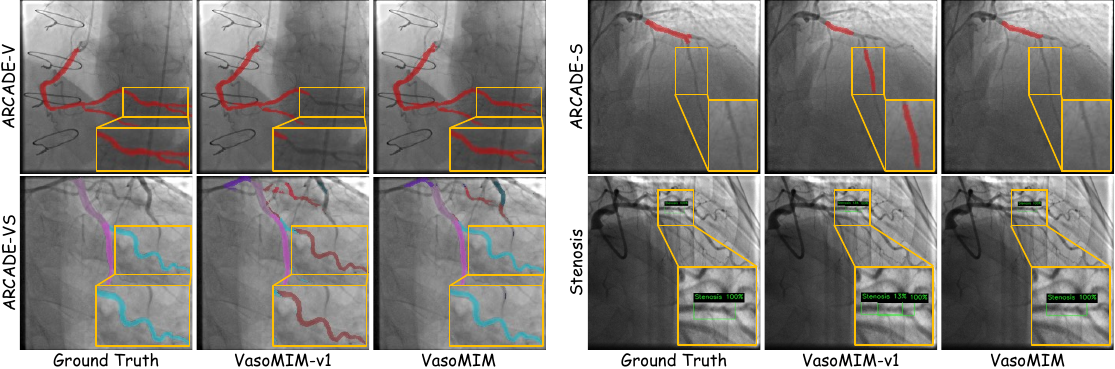}}
\caption{Qualitative comparison between VasoMIM-v1~\cite{huang2026vasomim} and VasoMIM on vessel segmentation (ARCADE-V), stenosis segmentation (ARCADE-S), vessel segment segmentation (ARCADE-VS), and stenosis detection (Stenosis). Yellow boxes highlight improved structural continuity, lesion delineation, or localization.}
\label{fig:visualization_results}
\end{figure*}

\subsection{Detailed Comparison with VasoMIM-v1~\cite{huang2026vasomim}}
\textbf{Instance-level \textit{vs.} Population-level Guidance.}  
We further compare instance-level and population-level anatomical guidance to clarify the methodological difference between VasoMIM-v1 and the current version, as shown in Table~\ref{table:guidance_comparison_aaai}. \textbf{\textit{i)}} Frangi-only guidance brings marginal improvements over random masking, with $+0.05\%$ DSC on ARCADE-V and $+0.01\%$ DSC on XACV. Moreover, its clDice slightly decreases on ARCADE-V. This suggests that although Frangi-only guidance is effective in VasoMIM-v1, its reliability becomes limited when pre-training is scaled to a larger and more diverse X-ray angiogram dataset with low contrast, bony structures, and imaging noise.\begin{figure}[htbp]
\centering\centerline{\includegraphics{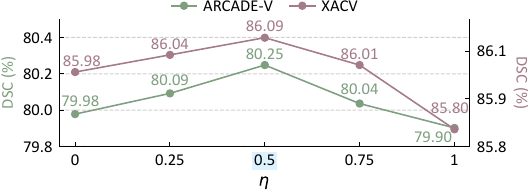}}
\caption{Hyper-parameter sensitivity of the mixing coefficient $\eta$ on ARCADE-V and XACV. A larger $\eta$ assigns a higher relative weight to Frangi-derived instance-level guidance, with $\eta=0$ and $\eta=1$ corresponding to segmentor-only and Frangi-only guidance, respectively. All results are averaged over five random seeds.}
\label{fig:eta}
\end{figure}\textbf{\textit{ii)}} In contrast, segmentor-only guidance provides more stable improvements across both datasets and evaluation metrics. These results indicate that population-level anatomical knowledge learned from large-scale data can provide more robust guidance than relying solely on instance-specific vesselness extraction. \textbf{\textit{iii)}} Furthermore, combining the two sources significantly outperforms either guidance source alone ($p<0.05$). Notably, the resulting gain over random masking exceeds the sum of the gains achieved by Frangi-only and segmentor-only guidance. A plausible explanation is that the two guidance sources provide complementary information. Importantly, although the two guidance maps are linearly combined in Eq.~(\ref{eq:mixing}), their downstream performance gains are not expected to be additive, since the resulting co-guidance is used to determine the patch-wise masking distribution and subsequently influences the nonlinear pre-training and fine-tuning processes. By integrating complementary instance-level and population-level anatomical information, co-guidance provides more reliable localization of vessel-relevant regions, as shown in Fig.~\ref{fig:segguidance} (Right), thereby enabling gains beyond those achieved by either guidance source independently.

\textbf{Effect of the Mixing Coefficient $\eta$.} We further investigate the relative contribution of the two guidance sources by varying $\eta$ in Eq.~(\ref{eq:mixing}) from $0$ to $1$. As shown in Fig.~\ref{fig:eta}, both ARCADE-V and XACV exhibit a consistent single-peaked trend. These results indicate that neither guidance source alone is optimal, while a balanced combination ($\eta=0.5$) provides the most effective anatomical guidance. The smooth peak around the default setting also supports that the co-guidance gain arises from the complementary integration of the two guidance sources rather than from a particular endpoint.

\textbf{Qualitative Visualization.} We further provide qualitative comparisons between VasoMIM-v1 and VasoMIM on four downstream tasks. As shown in Fig.~\ref{fig:visualization_results}, both methods can capture the major vascular structures, while VasoMIM generally produces predictions that are more consistent with the ground truth. Compared with VasoMIM-v1, the current version shows improved structural completeness for elongated vessels, clearer delineation of task-relevant regions, and more compact localization of stenotic lesions. These improvements are particularly meaningful for X-ray angiogram analysis, where small discontinuities, boundary deviations, or imprecise lesion localization may affect the interpretation of vascular morphology. Overall, these qualitative results suggest that the representations learned by VasoMIM transfer more effectively to downstream tasks.

Overall, these results show that the journal version addresses a concrete limitation of the Frangi-only guidance in VasoMIM-v1 by introducing population-level anatomical guidance and integrating it with instance-level vesselness cues. Importantly, this population-level guidance does not introduce additional modules or downstream inference cost compared with VasoMIM-v1. Since the segmentor has already been used for the anatomical consistency loss in VasoMIM-v1, the current version reuses its probability map to guide patch masking, thereby improving guidance reliability with essentially unchanged computational overhead.

\subsection{Masking Strategies}
\textbf{Effect of Patch Sampling Strategy.}  We first evaluate whether the improvement comes merely from identifying vessel-relevant patches or from the proposed patch-wise vascular anatomical distribution. As shown in Table~\ref{table:guidance_soft_hard}, replacing $f(g_i)$ with uniform sampling over guided patches degrades the performance. This indicates that simply locating vessel-relevant regions is not sufficient. Assigning higher sampling probabilities to patches with richer vascular content is also important for effective representation learning.

\begin{table}[htbp]
\caption{Ablation of patch sampling strategy and guidance format. DSC is reported in this table. ``Format'' refers to the segmentor output used in guidance: ``Hard'' denotes a binarized mask, ``Soft'' denotes a probability map, and ``$-$'' indicates that no segmentor output is used. Paired $t$-tests are conducted between each variant and VasoMIM using five seed-level results. $\ast$: $p<0.05$, ${\rm n.s.}$: not significant.}~\label{table:guidance_soft_hard}
\centering
\resizebox{\linewidth}{!}{
\begin{tabular}{lllll}
\toprule
Guidance & Format & Sampling & ARCADE-V & XACV \\ \midrule
$-$  & $-$ & Random & $79.85${\tiny$\pm0.14$}$^{\ast}$ & $85.79${\tiny$\pm0.52$}$^{\rm n.s.}$ \\
\multirow{6}{*}{Co-guidance} & \multirow{2}{*}{Soft} & \multirow{2}{*}{Uniform} & $79.99${\tiny$\pm0.14$}$^{\ast}$ & $85.94${\tiny$\pm0.06$}$^{\ast}$ \\
 & & & {\color{deepgreen}$\uparrow0.14$} & {\color{deepgreen}$\uparrow0.15$} \\
 & \multirow{2}{*}{Hard} & \multirow{2}{*}{$f(g_i)$} & $\underline{80.03}${\tiny$\pm0.04$}$^{\ast}$ & $\underline{85.98}${\tiny$\pm0.03$}$^{\ast}$ \\
 & & & {\color{deepgreen}$\uparrow0.18$} & {\color{deepgreen}$\uparrow0.19$} \\
 & \cellcolor{lightblue} & \cellcolor{lightblue} & \cellcolor{lightblue}$\bm{80.25}${\tiny$\pm0.12$} & \cellcolor{lightblue}$\bm{86.09}${\tiny$\pm0.09$} \\
 & \multirow{-2}{*}{\cellcolor{lightblue}Soft} & \multirow{-2}{*}{\cellcolor{lightblue}$f(g_i)$} & \cellcolor{lightblue}{\color{deepgreen}$\uparrow\bm{0.40}$} & \cellcolor{lightblue}{\color{deepgreen}$\uparrow\bm{0.30}$} \\ \bottomrule
\end{tabular}
}
\end{table}
\begin{table}[htbp]
\caption{Evaluation of different masking strategies. DSC is reported in this table.}
\label{table:strategy}
\centering
\resizebox{\linewidth}{!}{
\begin{tabular}{llllll}
\toprule
Case & Manner & $\beta_0$ & $\beta_E$ & ARCADE-V & XACV \\ \midrule
Random & $-$ & $0$ & $0$ & $79.85${\tiny$\pm0.14$} & $85.79${\tiny$\pm0.52$} \\ \cmidrule{2-6}
\multirow{10}{*}{Guided} & \multirow{8}{*}{Weak-to-Strong} & \cellcolor{lightblue} & \cellcolor{lightblue} & \cellcolor{lightblue}$\bm{80.25}${\tiny$\pm0.12$} & \cellcolor{lightblue}$\bm{86.09}${\tiny$\pm0.09$} \\
 & & \multirow{-2}{*}{\cellcolor{lightblue}$0$} & \multirow{-2}{*}{\cellcolor{lightblue}$0.5$} & \cellcolor{lightblue}{\color{deepgreen}$\uparrow\bm{0.40}$} & \cellcolor{lightblue}{\color{deepgreen}$\uparrow\bm{0.30}$} \\
 & & \multirow{2}{*}{$0$} & \multirow{2}{*}{$0.75$} & $79.91${\tiny$\pm0.13$} & $\underline{85.99}${\tiny$\pm0.07$} \\
 & & & & {\color{deepgreen}$\uparrow0.06$} & {\color{deepgreen}$\uparrow0.20$} \\
 & & \multirow{2}{*}{$0$} & \multirow{2}{*}{$1$} & $\underline{79.96}${\tiny$\pm0.06$} & $85.94${\tiny$\pm0.13$} \\
 & & & & {\color{deepgreen}$\uparrow0.11$} & {\color{deepgreen}$\uparrow0.15$} \\
 & & \multirow{2}{*}{$1$} & \multirow{2}{*}{$1$} & $78.04${\tiny$\pm0.12$} & $83.89${\tiny$\pm1.53$} \\
 & & & & {\color{red}$\downarrow1.81$} & {\color{red}$\downarrow1.90$} \\ \cmidrule{2-6}
 & \multirow{2}{*}{Strong-to-Weak} & \multirow{2}{*}{$0.5$} & \multirow{2}{*}{$0$} & $79.79${\tiny$\pm0.07$} & $85.63${\tiny$\pm0.34$} \\
 & & & & {\color{red}$\downarrow0.06$} & {\color{red}$\downarrow0.16$} \\
 \bottomrule
\end{tabular}
}
\end{table}

\textbf{Soft \textit{vs}. Hard Co-guidance.} We then study the influence of the guidance format, as provided in Table~\ref{table:guidance_soft_hard}. Compared with hard co-guidance, soft co-guidance yields additional gains of $+0.22\%$ and $+0.11\%$ DSC on ARCADE-V and XACV, respectively. This suggests that preserving the probability distribution of the segmentor output is more beneficial than binarizing it into a hard mask. A possible reason is that soft guidance retains uncertainty around faint vessels and ambiguous boundaries, thereby providing more informative anatomical cues for pre-training.

\textbf{Weak-to-Strong Design.} We further investigate the optimal scheduling of anatomical guidance by varying $\beta_0$ and $\beta_E$. From Table~\ref{table:strategy}, we observe that overly strong guidance, especially when applied from the beginning of pre-training, degrades performance. This result is intuitive since aggressive masking of too many vessel-containing patches eliminates the semantic context necessary for reconstruction, forcing the model to solve an ill-posed problem based solely on background noise. Furthermore, we evaluate a reversed strong-to-weak approach. This approach underperforms even the fully random baseline, underscoring the critical importance of the proposed weak-to-strong design, which progressively increases task difficulty to ensure stable representation learning.

\begin{figure}[htbp]
\centering\centerline{\includegraphics{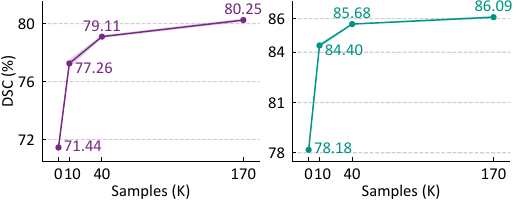}}
\caption{Data scaling analysis. We evaluate the impact of increasing pre-training data size from $10$K to $170$K on downstream performance. \textbf{Left:} ARCADE-V. \textbf{Right:} XACV.}
\label{fig:data_scaling_law}
\end{figure}
\begin{figure}[htbp]
\centering\centerline{\includegraphics{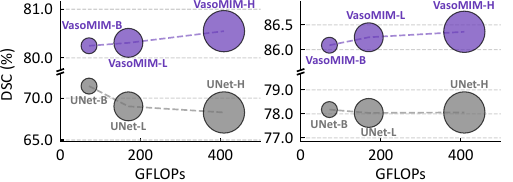}}
\caption{Scalability analysis of VasoMIM. We evaluate the impact of scaling the model from ViT-B to ViT-L and ViT-H on downstream performance. \textbf{Left:} ARCADE-V. \textbf{Right:} XACV.}
\label{fig:model_scaling_law}
\end{figure}

\subsection{Scalability Analysis}
We investigate the scaling capability of VasoMIM regarding both data scale and model capacity. \textbf{\textit{i)} Data Scaling.} We scale the pre-training dataset from $10$K to $170$K and evaluate downstream performance in Fig.~\ref{fig:data_scaling_law}. The performance gradually increases with more data. However, we find that the performance gain from $40$K to $170$K is less pronounced than the initial improvement from $10$K to $40$K. These findings are consistent with recent medical SSL studies, such as VoCo~\cite{wu2026large}. \textbf{\textit{ii)} Model Scaling\footnote{We release all pre-trained models on \href{https://huggingface.co/waha2000huang/VasoMIM}{Hugging Face}.}.} We further scale the backbone from ViT-B ($86$ M) to ViT-L ($307$ M) and ViT-H ($632$ M). As illustrated in Fig.~\ref{fig:model_scaling_law}, while increasing model capacity consistently yields higher DSC, the improvements are marginal relative to the substantial increase in computational cost.

We hypothesize three primary factors contributing to the above results: \textbf{\textit{i)} Architectural Bottlenecks.} In this study, we mainly focus on developing a domain-specific pre-training framework instead of designing novel architectures. The backbone decoder, \textit{i.e.}, UNet~\cite{ronneberger2015u}, may lack scalability. \textbf{\textit{ii)} Data Quality.} XA-$170$K inevitably contains variable image quality and artifacts, which may degrade representation learning. \textbf{\textit{iii)} Convergence.} Given that VasoMIM already delivers substantial improvements over the baseline, the performance may be approaching the ceiling under the current experimental protocol.

\begin{figure*}[t]
\centering\centerline{\includegraphics{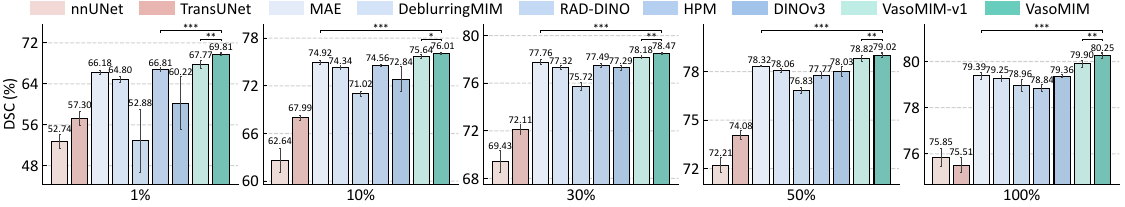}}
\caption{Label-efficient fine-tuning on ARCADE-V. VasoMIM consistently outperforms competing SSL alternatives, notably achieving superior performance with only $10\%$ of labels compared with nnUNet~\cite{isensee2021nnu} and TransUNet~\cite{chen2024transunet} trained from scratch using $100\%$ of labels. $\ast$: $p<0.05$, $\ast\ast$: $p<0.01$, $\ast\ast\ast$: $p<0.001$ (paired $t$-test).}
\label{fig:data_efficiency}
\end{figure*}

\subsection{Label Efficiency}
We compare VasoMIM against five leading SSL baselines, two strong models trained from scratch, and our previous VasoMIM-v1~\cite{huang2026vasomim} across varying label-scarce scenarios. As illustrated in Fig.~\ref{fig:data_efficiency}, VasoMIM consistently outperforms all evaluated methods at every data scale. \textit{\textbf{i)}} Compared with models trained from scratch, VasoMIM demonstrates clear advantages in label-scarce scenarios. Utilizing merely $1\%$ of the training data, VasoMIM achieves a DSC of $69.81\%$, exceeding nnUNet~\cite{isensee2021nnu} and TransUNet~\cite{chen2024transunet} by $+17.07\%$ and $+12.51\%$, respectively.\begin{table}[htbp]
\caption{Impact of pre-training dataset. We compare models pre-trained on generic, medical-specific, and our XA-$170$K datasets. DSC is reported in this table.}
\label{table:pt_dataset}
\centering
\resizebox{\linewidth}{!}{
\begin{tabular}{llll}
\toprule
Method & Pre-training Data & ARCADE-V & XACV \\ \midrule
\multirow{3}{*}{MAE~\cite{he2022masked} {\tiny\color{gray}[\textit{CVPR}'22]}} & ImageNet-$1$K & $78.10${\tiny$\pm0.21$} & $83.69${\tiny$\pm1.48$} \\
 & XA-$170$K & $79.39${\tiny$\pm0.15$} & $84.84${\tiny$\pm1.79$} \\
 & $\Delta$ & {\color{deepgreen}$\uparrow1.29$} & {\color{deepgreen}$\uparrow1.15$} \\ 
 \midrule
\multirow{3}{*}{LocalMIM~\cite{wang2023masked} {\tiny\color{gray}[\textit{CVPR}'23]}} & ImageNet-$1$K & $77.55${\tiny$\pm0.66$} & $83.00${\tiny$\pm0.61$} \\
 & XA-$170$K & $78.79${\tiny$\pm0.20$} & $83.18${\tiny$\pm1.84$} \\ 
 & $\Delta$ & {\color{deepgreen}$\uparrow1.24$} & {\color{deepgreen}$\uparrow0.18$}  \\
 \midrule
\multirow{3}{*}{DeblurringMIM~\cite{kang2024deblurring} {\tiny\color{gray}[\textit{MedIA}'24]}} & US-Esaote-$280$K & $77.80${\tiny$\pm0.24$} & $84.46${\tiny$\pm0.10$} \\
 & XA-$170$K & $79.25${\tiny$\pm0.13$} & $85.38${\tiny$\pm0.45$} \\ 
 & $\Delta$ & {\color{deepgreen}$\uparrow1.45$} & {\color{deepgreen}$\uparrow0.92$} \\
 \midrule
\multirow{3}{*}{CheXWorld~\cite{yue2025chexworld} {\tiny\color{gray}[\textit{CVPR}'25]}} & X-rays-$500$K & $77.12${\tiny$\pm0.37$} & $82.12${\tiny$\pm1.10$} \\
 & XA-$170$K & $78.18${\tiny$\pm0.24$} & $83.81${\tiny$\pm0.79$} \\ 
 & $\Delta$ & {\color{deepgreen}$\uparrow1.06$} & {\color{deepgreen}$\uparrow1.69$} \\
 \midrule
\multirow{3}{*}{HPM~\cite{wang2025bootstrap} {\tiny\color{gray}[\textit{TPAMI}'25]}} & ImageNet-$1$K & $78.45${\tiny$\pm0.06$} & $84.82${\tiny$\pm0.16$} \\
 & XA-$170$K & $78.84${\tiny$\pm0.15$} & $85.19${\tiny$\pm0.07$} \\ 
 & $\Delta$ & {\color{deepgreen}$\uparrow0.39$} & {\color{deepgreen}$\uparrow0.37$} \\
\bottomrule
\end{tabular}
}
\end{table}Remarkably, when fine-tuned on only $10\%$ of the labeled data, VasoMIM equipped with a standard UNet~\cite{ronneberger2015u} backbone attains $76.01\%$ DSC. This performance surpasses that of nnUNet and TransUNet trained from scratch using the entire $100\%$ labeled dataset by $+0.16\%$ and $+0.50\%$, respectively. These results indicate that self-supervised pre-training can substantially reduce the demand for manual annotations. \textit{\textbf{ii)}} VasoMIM also consistently outperforms the SSL baselines across different label proportions. The advantage is particularly pronounced in the extreme low-data regime. For instance, under the $1\%$ setting, VasoMIM surpasses the strongest SSL competitor, HPM~\cite{wang2025bootstrap}, by $+3.00\%$ DSC. This superiority remains consistent across the $10\%$, $30\%$, $50\%$, and $100\%$ label partitions, demonstrating the effectiveness of incorporating vascular anatomical knowledge into self-supervised pre-training. \textit{\textbf{iii)}} Compared with our previous VasoMIM-v1~\cite{huang2026vasomim}, VasoMIM yields consistent improvements under different label ratios. Notably, the gain is especially evident under the $1\%$ setting, where VasoMIM improves DSC from $67.77\%$ to $69.81\%$. This suggests that VasoMIM is not merely beneficial in the full-data setting, but also improves the robustness when annotations are scarce. Overall, these results underscore the value of VasoMIM for label-efficient X-ray angiogram analysis.

\begin{figure}[htbp]
\centering\centerline{\includegraphics{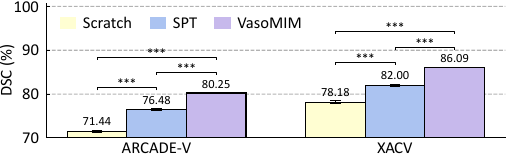}}
\caption{Segmentor-based Pre-training (SPT) \textit{vs.} MIM. VasoMIM significantly outperforms SPT, achieving DSC improvements of $+3.77\%$ on ARCADE-V and $+4.09\%$ on XACV. $\ast\ast\ast$: $p<0.001$ (paired $t$-test).}
\label{fig:spt}
\end{figure}

\subsection{Effectiveness of Domain-specific Pre-training Dataset}
Five representative SSL methods are selected to validate the contribution of XA-$170$K. Specifically, three generic vision baselines~\cite{he2022masked,wang2023masked,wang2025bootstrap} pre-trained on ImageNet-$1$K~\cite{deng2009imagenet} and two medical-domain methods~\cite{kang2024deblurring,yue2025chexworld} pre-trained on ultrasound and chest X-ray data are evaluated. Their officially released checkpoints are utilized for fine-tuning. As demonstrated in Table~\ref{table:pt_dataset}, pre-training on XA-$170$K consistently yields better performance on X-ray angiogram analysis tasks than pre-training on generic natural images or other medical imaging modalities. These improvements reveal a clear domain gap between existing generic/medical pre-training datasets and X-ray angiograms, highlighting the necessity of XA-$170$K for domain-specific vascular representation learning.


\subsection{Segmentor-based Pre-training \textit{vs.} MIM}
To verify that the performance gains stem from the proposed pre-training framework rather than merely from exposure to additional vessel-related pseudo-labels, VasoMIM is compared against a segmentor-based pre-training (SPT) baseline. Specifically, UNet is pre-trained in a fully supervised manner using pseudo-labels generated by Frangi filter on XA-$170$K. As illustrated in Fig.~\ref{fig:spt}, while SPT provides a significant improvement over training from scratch, it consistently lags behind VasoMIM.  Notably, VasoMIM achieves significant gains of $+3.77\%$ DSC on ARCADE-V and $+4.09\%$ DSC on XACV compared to the SPT baseline. This considerable performance gap indicates that SPT may be limited by the noise and artifacts present in the Frangi-generated pseudo-labels, whereas the MIM pre-training paradigm, \textit{i.e.}, VasoMIM, successfully extracts more robust and transferable vascular representations.

\begin{figure}[htbp]
\centering\centerline{\includegraphics{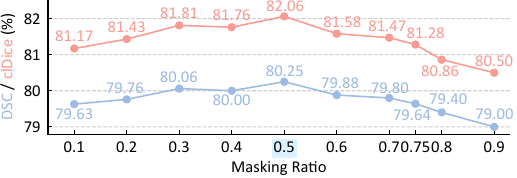}}
\caption{Hyper-parameter sensitivity of the masking ratio $\gamma$ on ARCADE-V. All results are averaged over five random seeds.}
\label{fig:masking_ratio_arcadev}
\end{figure}
\begin{table}[htbp]
\caption{Sensitivity analysis of VasoMIM to the percentile threshold $\alpha$ of Frangi filter.}\label{table:frangi_quality}
\centering
\resizebox{\linewidth}{!}{
    \begin{tabular}{llllll}
    \toprule
    \multirow{2}{*}{$\alpha$} & \multirow{2}{*}{Method} & \multicolumn{2}{c}{ARCADE-V} & \multicolumn{2}{c}{XACV} \\ \cmidrule{3-6} 
     & & DSC (\%) & clDice (\%) & DSC (\%) & clDice (\%) \\ \midrule
    \multirow{2}{*}{$50$} & Frangi & $12.06$ & $8.20$ & $32.16$ & $22.58$ \\
     & VasoMIM & $79.71${\tiny$\pm0.12$} & $81.26${\tiny$\pm0.21$} & $86.02${\tiny$\pm0.08$} & $84.05${\tiny$\pm0.12$} \\ \midrule
    \multirow{2}{*}{$80$} & Frangi & $27.06$ & $22.34$ & $66.70$ & $62.51$ \\
     & VasoMIM & $79.67${\tiny$\pm0.18$} & $81.31${\tiny$\pm0.22$} & $86.03${\tiny$\pm0.06$} & $84.03${\tiny$\pm0.18$} \\ \midrule
    \multirow{2}{*}{$92$} & Frangi & $41.30$ & $40.91$ & $58.46$ & $57.15$ \\
     & \cellcolor{lightblue}VasoMIM & \cellcolor{lightblue}$80.25${\tiny$\pm0.12$} & \cellcolor{lightblue}$82.06${\tiny$\pm0.18$}  & \cellcolor{lightblue}$86.09${\tiny$\pm0.09$} & \cellcolor{lightblue}$84.12${\tiny$\pm0.17$} \\ 
    \bottomrule
    \end{tabular}
    }
\end{table}

\subsection{Hyper-parameter Analysis}
\textbf{Masking Ratio.} Fig.~\ref{fig:masking_ratio_arcadev} illustrates the impact of varying the masking ratio $\gamma$ on ARCADE-V. We observe that a moderate masking ratio (\textit{i.e.}, $\gamma=0.50$) yields optimal results. This finding differs from common practice in the general vision domain, where higher ratios (\textit{e.g.}, $\gamma=0.75$ in MAE~\cite{he2022masked}) are typically preferred. We attribute this discrepancy to the intrinsic characteristics of X-ray angiograms. Compared with natural images, X-ray angiograms are dominated by background regions and contain only sparse vascular structures. Consequently, excessively high masking ratios tend to obscure the sparse anatomical cues essential for reconstruction. This forces the model to focus on reconstructing non-informative background noise rather than learning vascular representations.

\textbf{Sensitivity to Percentile Threshold.} We investigate the sensitivity of VasoMIM to the percentile threshold $\alpha$ of Frangi filter. For each value of $\alpha$, we regenerate the corresponding pseudo-labels on XA-$170$K, retrain the lightweight segmentor, and reconstruct the co-guidance for pre-training. As detailed in Table~\ref{table:frangi_quality}, the segmentation performance of Frangi filter fluctuates drastically depending on the percentile threshold $\alpha$.\footnote{Here, the DSC of Frangi filter is reported as a descriptive measure of the changes in extracted vascular structures rather than a measure of anatomical guidance quality.} The downstream effect of $\alpha$ is dataset-dependent. ARCADE-V mainly annotates major vessels.\begin{table}[htbp]
\caption{Evaluation of different filters. $\dagger$: General edge detector. $\ddagger$: Vesselness filter. $-$ denotes the vanilla baseline, \textit{i.e.,} MAE ($\gamma=0.5$), without anatomy-guided masking or anatomical consistency loss.}\label{table:vesselness_filter}
\centering
\resizebox{\linewidth}{!}{
\begin{tabular}{lllll}
\toprule
\multirow{2}{*}{Filter} & \multicolumn{2}{c}{ARCADE-V} & \multicolumn{2}{c}{XACV} \\ \cmidrule{2-5} 
 & DSC (\%) & clDice (\%) & DSC (\%) & clDice (\%) \\ \midrule
$-$ & $79.31${\tiny$\pm0.16$} & $81.05${\tiny$\pm0.25$} & $84.52${\tiny$\pm2.12$} & $82.25${\tiny$\pm2.32$} \\
\multirow{2}{*}{Canny$^\dagger$~\cite{canny1986computational} {\tiny\color{gray}[\textit{TPAMI}'86]}} & $79.32${\tiny$\pm0.12$} & $80.86${\tiny$\pm0.24$} & $85.73${\tiny$\pm0.06$} & $83.60${\tiny$\pm0.09$} \\
 & {\color{deepgreen}$\uparrow0.01$} & {\color{red}$\downarrow0.19$} & {\color{deepgreen}$\uparrow1.21$} & {\color{deepgreen}$\uparrow1.35$} \\
\multirow{2}{*}{Sato$^\ddagger$~\cite{sato1998three} {\tiny\color{gray}[\textit{MedIA}'98]}} & $\underline{79.99}${\tiny$\pm0.14$} & $81.53${\tiny$\pm0.31$} & $85.84${\tiny$\pm0.09$} & $83.75${\tiny$\pm0.09$} \\
 & ${\color{deepgreen}\uparrow0.68}$ & {\color{deepgreen}$\uparrow0.48$} & {\color{deepgreen}$\uparrow1.32$} & {\color{deepgreen}$\uparrow1.50$} \\
\multirow{2}{*}{Meijering$^\ddagger$~\cite{meijering2004design} {\tiny\color{gray}[\textit{Cytometry A}'04]}} & $79.96${\tiny$\pm0.06$} & $\underline{81.63}${\tiny$\pm0.12$} & $\underline{85.95}${\tiny$\pm0.05$} & $\underline{84.00}${\tiny$\pm0.05$} \\
 & {\color{deepgreen}$\uparrow0.65$} & {\color{deepgreen}$\uparrow0.58$} & {\color{deepgreen}$\uparrow1.43$} & {\color{deepgreen}$\uparrow1.75$} \\
\cellcolor{lightblue} & \cellcolor{lightblue}$\bm{80.25}${\tiny$\pm0.12$} & \cellcolor{lightblue}$\bm{82.06}${\tiny$\pm0.18$} & \cellcolor{lightblue}$\bm{86.09}${\tiny$\pm0.09$} & \cellcolor{lightblue}$\bm{84.12}${\tiny$\pm0.17$} \\
\cellcolor{lightblue}\multirow{-2}{*}{\cellcolor{lightblue}Frangi$^\ddagger$~\cite{frangi1998multiscale} {\tiny\color{gray}[\textit{MICCAI}'98]}} & \cellcolor{lightblue}{\color{deepgreen}$\uparrow\bm{0.94}$} & \cellcolor{lightblue}{\color{deepgreen}$\uparrow\bm{1.01}$} & \cellcolor{lightblue}{\color{deepgreen}$\uparrow\bm{1.57}$} & \cellcolor{lightblue}{\color{deepgreen}$\uparrow\bm{1.87}$} \\ \bottomrule
\end{tabular}
}
\end{table}Therefore, lowering $\alpha$ introduces additional weak responses and fine vascular structures that are less relevant to its downstream target, reducing the selectivity of anatomical guidance.\begin{table}[htbp]
\caption{Evaluation of different metric functions.}\label{table:metric_function}
\centering
\begin{tabular}{lllll}
\toprule
\multirow{2}{*}{$\mathcal{L}$} & \multicolumn{2}{c}{ARCADE-V} & \multicolumn{2}{c}{XACV} \\ \cmidrule{2-5} 
 & DSC (\%) & clDice (\%) & DSC (\%) & clDice (\%) \\ \midrule
$-$ & $79.87${\tiny$\pm0.09$} & $81.41${\tiny$\pm0.11$} & $85.92${\tiny$\pm0.04$} & $83.78${\tiny$\pm0.15$} \\
\multirow{2}{*}{L$_1$} & $79.64${\tiny$\pm0.17$} & $81.21${\tiny$\pm0.24$} & $85.87${\tiny$\pm0.08$} & $83.85${\tiny$\pm0.07$} \\
 & {\color{red}$\downarrow0.23$} & {\color{red}$\downarrow0.20$} & {\color{red}$\downarrow0.05$} & {\color{deepgreen}$\uparrow0.07$} \\
\multirow{2}{*}{Dice} & $79.94${\tiny$\pm0.11$} & $81.64${\tiny$\pm0.19$} & $85.93${\tiny$\pm0.08$} & $84.04${\tiny$\pm0.14$} \\
 & {\color{deepgreen}$\uparrow0.07$} & {\color{deepgreen}$\uparrow0.23$} & {\color{deepgreen}$\uparrow0.01$} & {\color{deepgreen}$\uparrow0.26$} \\
\multirow{2}{*}{Dice$+$CE} & $\underline{80.06}${\tiny$\pm0.07$} & $\underline{81.80}${\tiny$\pm0.08$} & $\underline{86.08}${\tiny$\pm0.09$} & $\underline{84.09}${\tiny$\pm0.10$} \\
 & {\color{deepgreen}$\uparrow0.19$} & {\color{deepgreen}$\uparrow0.39$} & {\color{deepgreen}$\uparrow0.16$} & {\color{deepgreen}$\uparrow0.31$} \\
\cellcolor{lightblue} & \cellcolor{lightblue}$\bm{80.25}${\tiny$\pm0.12$} & \cellcolor{lightblue}$\bm{82.06}${\tiny$\pm0.18$} & \cellcolor{lightblue}$\bm{86.09}${\tiny$\pm0.09$} & \cellcolor{lightblue}$\bm{84.12}${\tiny$\pm0.17$} \\
\cellcolor{lightblue}\multirow{-2}{*}{CE} & \cellcolor{lightblue}{\color{deepgreen}$\uparrow\bm{0.38}$} &\cellcolor{lightblue}{\color{deepgreen}$\uparrow\bm{0.65}$} & \cellcolor{lightblue}{\color{deepgreen}$\uparrow\bm{0.17}$} & \cellcolor{lightblue}{\color{deepgreen}$\uparrow\bm{0.34}$} \\ \bottomrule
\end{tabular}
\end{table}Consequently, ARCADE-V shows a clear preference for the more selective setting of $\alpha=92$. In contrast, XACV contains more complete annotations of thin and distal vessels. Lowering $\alpha$ therefore recovers additional fine vascular structures that remain informative for XACV, while simultaneously introducing weaker and noisier responses. The increased anatomical completeness thus partially compensates for the reduced specificity, resulting in relatively stable downstream performance across different values of $\alpha$.

\textbf{Different Filters.} We evaluate whether the effectiveness of VasoMIM is sensitive to the choice of the anatomical extractor. To this end, we replace Frangi filter with one general edge detector~\cite{canny1986computational} and two representative vesselness filters~\cite{sato1998three,meijering2004design}. For each alternative extractor, we use it to generate pseudo-labels on XA-$170$K, retrain the lightweight segmentor, and reconstruct the corresponding co-guidance map for pre-training. As shown in Table~\ref{table:vesselness_filter}, the vesselness-based variants consistently outperform the anatomy-agnostic baseline on both ARCADE-V and XACV, suggesting that the proposed anatomy-aware mechanism can generalize across different vesselness extractors. In comparison, Canny provides less consistent benefits, suggesting that vessel-specific structural responses are more suitable for anatomical guidance than generic edge information. Among the evaluated extractors, Frangi filter achieves the best overall performance, supporting its use as the default anatomical extractor.

\textbf{Different Metric Functions $\mathcal{L}$.} We further investigate the choice of the metric function $\mathcal{L}$ in the anatomical consistency loss. As shown in Table~\ref{table:metric_function}, L$_1$ loss slightly degrades performance, indicating that point-wise regression is insufficient to enforce anatomical consistency. In contrast, Dice loss improves the baseline, suggesting that region-level overlap is useful for preserving vascular anatomy. Combining Dice loss with CE loss further improves the performance. Among all metric functions, the default CE loss achieves the best overall performance. We attribute this to the fact that CE loss provides dense supervision between the vascular probability maps of the original and reconstructed X-ray angiograms, which is well aligned with our goal of enforcing anatomical consistency during pre-training. Therefore, we adopt CE loss as the default metric function.

\section{Limitations and Future Works} \label{sec:limitation}
Although VasoMIM demonstrates strong empirical performance across diverse X-ray angiogram analysis tasks, it has several limitations that can be further explored in future works. First, the scalability of the model is currently bottlenecked by the UNet decoder. As the primary focus of this study lies in establishing a domain-specific pre-training framework rather than architectural design, future work will explore the integration of more advanced, scalable network architectures to fully unlock the potential of large-scale pre-training. Second, XA-$170$K is restricted to coronary arteries. This limitation arises from the severe scarcity of publicly available X-ray angiogram datasets for other anatomical regions, such as the aorta, cerebral vessels, and peripheral vessels. To address this gap, we plan to collaborate with clinical institutions to curate a comprehensive, multi-region dataset. Third, VasoMIM currently derives its representations exclusively from visual features via SSL. Because angiograms in clinical practice are often accompanied by detailed radiological reports, extending this framework to incorporate vision-language pre-training, \textit{e.g.}, SigLIP 2~\cite{tschannen2025siglip}, will be a crucial next step. Finally, while we have comprehensively validated VasoMIM on core X-ray angiogram analysis tasks, its ultimate value should be further assessed in clinically oriented applications. Future efforts will focus on deploying these pre-trained representations for higher-level clinical assessments, such as the automated calculation of the SYNTAX score~\cite{head2014syntax}, to provide end-to-end clinical decision support.

\section{Conclusion} \label{sec:conclusion}
In this paper, we present \textbf{VasoMIM}, a \underline{\textbf{Vas}}cular anat\underline{\textbf{o}}my-aware \underline{\textbf{M}}asked \underline{\textbf{I}}mage \underline{\textbf{M}}odeling framework for large-scale X-ray angiogram pre-training. By explicitly integrating vascular anatomical guidance into the masking strategy, VasoMIM encourages the model to focus on sparse, semantically critical regions, thereby improving the learning of vascular representations. Furthermore, the proposed anatomical consistency loss enhances the discriminability of the learned representations. To facilitate large-scale pre-training, we have curated XA-$170$K, the largest X-ray angiogram dataset to date. Extensive experiments across six datasets, covering four distinct X-ray angiogram analysis tasks, demonstrate that VasoMIM achieves state-of-the-art performance. We believe this work offers a novel perspective on integrating anatomical knowledge into self-supervised learning, paving the way for the development of X-ray angiogram foundation models.

\bibliographystyle{IEEEtran}
\bibliography{reference}

\end{document}